\documentclass[lettersize,journal]{IEEEtran}
\usepackage{amsmath,amsfonts}
\usepackage{algorithmic}
\usepackage{algorithm}
\usepackage{array}
\usepackage[caption=false,font=normalsize,labelfont=sf,textfont=sf]{subfig}
\usepackage{textcomp}
\usepackage{stfloats}
\usepackage{url}
\usepackage{verbatim}
\usepackage{graphicx}
\usepackage{cite}
\newtheorem{proposition}{Proposition} 
\newtheorem{lemma}{Lemma}
\newtheorem{theorem}{Theorem}

\usepackage{tcolorbox}
\usepackage{booktabs}
\usepackage[table,xcdraw,dvipsnames]{xcolor}
\usepackage{tabularx}
\usepackage{multirow}
\usepackage[normalem]{ulem}
\usepackage{tikz}
\usepackage{empheq}
\definecolor{revisecolor}{RGB}{0,0,0}
\definecolor{commentcolor}{RGB}{102,153,204}
\usepackage{setspace}
\usepackage{float}
\usepackage{hyperref} 

\makeatletter
\newcommand{\Statex}[1][0pt]{\vspace{#1}}
\makeatother

\newtcolorbox{probox}[1][]{colback=black!5!white, colframe=black!70!black, boxsep=-4pt, grow to left by=4pt, left=10pt, grow to right by=4pt, right=10pt, top=10pt, bottom=10pt, #1}
\newtcolorbox{thmbox}[1][]{colback=thm!5!white,colframe=thm!60!black,boxsep=-4pt,grow to left by=4pt,left=10pt,grow to right by=4pt,right=10pt,top=10pt,bottom=10pt,#1}
\newtcolorbox{lembox}[1][]{colback=def!5!white,colframe=def!60!black,boxsep=-4pt,grow to left by=4pt,left=10pt,grow to right by=4pt,right=10pt,top=10pt,bottom=10pt,#1}

\begin{document}
\newcolumntype{C}{>{\centering\arraybackslash}X}

\definecolor{lightblue}{RGB}{91, 155, 213}
\definecolor{deepblue}{RGB}{68, 114, 196}
\definecolor{mygray}{RGB}{121, 121, 121}

\definecolor{forward}{RGB}{84, 130, 53}
\definecolor{inverse}{RGB}{47, 85, 151}
\definecolor{resist}{RGB}{128, 0, 128}
\definecolor{rebound}{RGB}{133, 19, 33}

\definecolor{def}{RGB}{119, 228, 200}
\definecolor{thm}{RGB}{69, 53, 193}

\colorlet{StepA}{RoyalBlue!70!Black}
\colorlet{StepB}{ForestGreen!60!Black}
\colorlet{StepC}{Plum!60!Black}
\colorlet{StepD}{BrickRed!65!Black}
 
\newcommand{\tv}{\Delta \theta}
\newcommand{\tf}[3][]{\Delta h_{#2}^{#3}(#1)}
\newcommand{\eg}{\textit{e.g.}}
\newcommand{\ie}{\textit{i.e.}}

\title{MAGIC: Achieving Superior Model Merging \\via Magnitude Calibration}

\author{
    Yayuan Li, Jian Zhang, Jintao Guo, Zihan Cheng, Lei Qi, Yinghuan Shi\textsuperscript{*}, Yang Gao
    \thanks{Yayuan Li, Jian Zhang, Jintao Guo, Yinghuan Shi and Yang Gao are with the State Key Laboratory for Novel Software Technology and the National Institute of Healthcare Data Science, Nanjing University, Nanjing, Jiangsu 210093, China. (e-mail: liyayuan@smail.nju.edu.cn; zhang.jian@nju.edu.cn; guojintao@smail.nju.edu.cn; syh@nju.edu.cn; gaoy@nju.edu.cn)}
    \thanks{Zihan Cheng is with Shanghai Jiao Tong University Medical School Affiliated Ruijin Hospital, Shanghai 200025, China; and is also with the National Institute of Healthcare Data Science, Nanjing University, Nanjing, Jiangsu 210093, China. (e-mail: chengzihan@sjtu.edu.cn).}
    \thanks{Lei Qi is with the School of Computer Science and Engineering, Key Lab of Computer Network and Information Integration, Southeast University, Nanjing, Jiangsu 211189, China. (e-mail: qilei@seu.edu.cn).}
    \thanks{The corresponding author of this work is Yinghuan Shi.}
}



\maketitle

\begin{abstract}

The proliferation of pre-trained models has given rise to a wide array of specialised, fine-tuned models. Model merging aims to merge the distinct capabilities of these specialised models into a unified model, requiring minimal or even no additional training.
A core objective of model merging is to ensure the merged model retains the behavioural characteristics of the specialised models, typically achieved through feature alignment. We identify that features consist of two critical components: direction and magnitude. 
Prior research has predominantly focused on directional alignment, while the influence of magnitude remains largely neglected, despite its pronounced vulnerability to perturbations introduced by common merging operations (\eg, parameter fusion and sparsification). 
Such perturbations to magnitude inevitably lead to feature deviations in the merged model from the specialised models, resulting in subsequent performance degradation.
To address this, we propose MAGnItude Calibration (MAGIC), a plug-and-play framework that rectifies layer-wise magnitudes in feature and weight spaces, with three variants. Specifically, our Feature Space Calibration (FSC) realigns the merged model's features using a small set of unlabelled data, while Weight Space Calibration (WSC) extends this calibration to the weight space without requiring additional data. Combining these yields Dual Space Calibration (DSC). Comprehensive experiments demonstrate that MAGIC consistently boosts performance across diverse Computer Vision tasks (+4.3\% on eight datasets) and NLP tasks (+8.0\% on Llama) without additional training. Our code is available at:
https://github.com/lyymuwu/MAGIC
\end{abstract}


\section{Introduction}
\label{sec:intro}

\IEEEPARstart{A}{s} deep learning flourishes, abundant fine-tuned models have been open-sourced on the Internet (\textit{e.g.}, HuggingFace and Civitai) over the past few years \cite{Goddard_2024_Arcees, Yang_2024_Model, Yang_2017_Deep, 9447164}. 
Since models specialised in different tasks may come from the same pre-trained model (\textit{e.g.}, Llama~\cite{touvron2023llama} or CLIP~\cite{radford2021learning}), an interesting question arises whether there is a way to directly merge task-specific knowledge into a single model without introducing extra parameters and retraining costs. 
Therefore, we can combine different styles of a generation model into a mixed style \cite{Nair_2024_MaxFusion, Shah_2025_ZipLoRA} or merge specialised models into a single model with multi-task processing ability \cite{Ilharco_2022_Editinga, Ortiz-Jimenez_2023_Task}.

\begin{figure}[!t]
    \centering
    \includegraphics[width=1\linewidth]{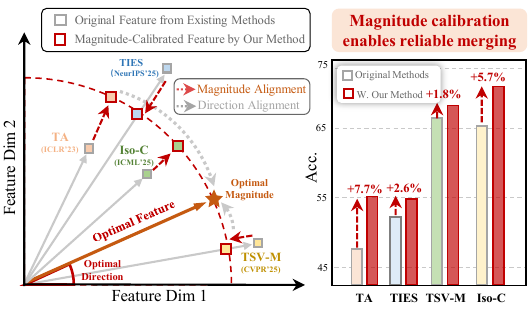}
    \caption{
        Current merging methods seek to approximate specialised models, whose features serve as the optimal reference. While features consist of both direction and magnitude, existing approaches overlook magnitude alignment. Our calibration method improves performance consistently by aligning feature magnitudes more closely with the optimal.
    }

    \label{fig:magnitude_matters}
\end{figure}
\begin{figure*}[!t]
    \centering
    \includegraphics[width=1\linewidth]{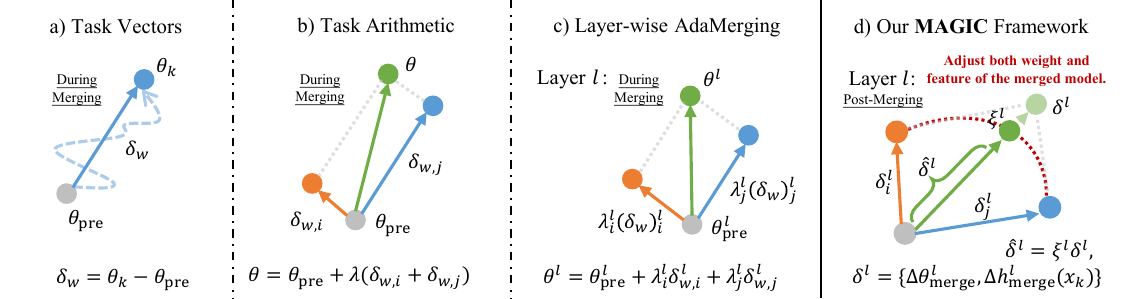}
    \caption{An illustration of task vectors based model merging methods. (a) Definition of task vectors, where $\theta_\text{pre}$ is the parameter of a pre-trained model, and $\theta_k$ is the model's weight fine-tuned on the $k$-th task. (b) Task Arithmetic \cite{Ilharco_2022_Editinga} merges models via the arithmetic mean of task vectors. (c) Layer-wise AdaMerging \cite{Yang_2024_AdaMerging} uses unlabelled data to learn different merge coefficients $\lambda_k^l$ for the task vector of each task $k$ and layer $l$. (d) Our method applies post-calibration after model merging, using adaptive estimation calibration coefficients $\xi^l$ to adjust both the task vector $\tv_\mathrm{merge}^l$ and task feature $\tf[x_k]{\mathrm{merge}}{l}$ at each layer.}
    \label{fig:operations}
\end{figure*}

Fortunately, model merging \cite{Ainsworth_2023_Git, Xu_2024_Training-Free} serves as a promising method for integrating the weights of task-specialised models into a single model.
A prior work, Weight Averaging~\cite{Izmailov_2019_Averaging}, achieves this by simply averaging their weights. 
In contrast, Task Arithmetic (TA)~\cite{Ilharco_2022_Editinga} introduces \emph{task vectors}, defined as the difference between fine-tuned and pre-trained weights, to isolate task-specific knowledge. 
This formulation not only provides an easier way to manipulate the contribution of each specialised model, but also discovers the inherent problem of model merging, where the conflicts of specialised models’ task vectors significantly impair the merged model performance.
Subsequent works~\cite{Yadav_2023_TIES-Merging, Du_2025_Neural, Gargiulo_2024_Task, Yu_2024_Language} try to restore the specialised model performance of merged models by aligning their intermediate feature directions with those of the specialised models.

However, they overlook another equally critical aspect: feature magnitude.
Through theoretical and experimental analysis, we observe that when only aligning the directions, various merging operations (\textit{e.g.}, parameter fusion or sparsification), inevitably make the feature magnitudes of the merged model deviate from those of the specialised model with significant performance degradation. 

As shown in Fig.~\ref{fig:magnitude_matters}, features from an intermediate layer of the merged model are projected onto two dimensions. A recent method, \eg, Iso-C~\cite{Marczak_2025_No} yields a smaller angle with the optimal feature (\ie, feature of the specialised model) compared to the conventional method, \eg, TA~\cite{Ilharco_2022_Editinga}, thereby achieving stronger directional alignment.
However, features from the merged model generally exhibit significant \textbf{``Magnitude Deviation''} from the optimal feature and fail to achieve perfect alignment with the optimal. 
This insight motivates our proposed plug-and-play MAGnItude Calibration (MAGIC) framework. Specifically, we develop MAGIC in both feature and weight space with three variants. Our Feature Space Calibration (FSC) method preserves feature direction while rescaling the deviated magnitude of the features induced by the task vector in the merged model to match those of the specialised models. This enables the merged model to achieve performance comparable to the specialised models.

Since FSC requires additional input data to calculate features and the magnitude variation between weights and features exhibits a linear correlation, the method is naturally extended to data-free Weight Space Calibration (WSC), which rescales the task vector's magnitude. Finally, Dual Space Calibration (DSC) is introduced to rescale in both spaces.

Furthermore, it is observed that different layers exhibit varying degrees of sensitivity to rescaling-based calibration. Certain layers that reside in sharp minima of the loss landscape are extremely sensitive to naive rescaling, particularly to scaling up, which leads to considerable performance degradation. These extremely sensitive layers are referred to as \textbf{``Magnitude-Sensitive Layers''}. 
To enhance the robustness of calibration, additional constraints are incorporated to suppress these magnitude-sensitive layers, thereby 
improving merging performance. The main research contributions are as follows:

\begin{itemize}
    \item Theoretical and empirical analyses expose severe magnitude deviation in merged models, clarify its impact on performance, and motivate effective mitigation strategies.
    
    \item A simple yet efficient post-calibration framework, MAGIC, is proposed with three variants in weight space (WSC), feature space (FSC) and dual space (DSC) to address magnitude deviation from the specialised models.
    
    \item Our proposed \textit{training-free} and \textit{plug-and-play} post-merging strategies demonstrate significant performance improvements across diverse scenarios (\textit{e.g.}, \textbf{8\%} improvement on Llama-2 for SOTA methods).
\end{itemize}


\section{Revisit Model Merging} 
\label{sec:model_merge}

Model merging exhibits promising performance in multi-task learning (MTL). 
Nevertheless, previous studies insufficiently addressed the issue of magnitude. 
This section first presents the necessary notation and outlines the model merging process in Section~\ref{sec:preliminary}, then analyzes the factors contributing to magnitude deviation in Section~\ref{sec:unstable_oper}, and theoretically examines the impact of magnitude deviation on performance as well as mitigation strategies in Section~\ref{sec:Magnitude Influence Performance} and Section~\ref{sec:Mitigate Magnitude Deviation}.

\subsection{Preliminary}
\label{sec:preliminary}

\textbf{Problem Setting}.
Let $\theta_{\mathrm{pre}}$ denote the parameters of a pre-trained model, and consider $K$ fine-tuned models derived from it. 
Each fine-tuned model consists of up to $L$ layers, where $\theta_k^l$ represents the parameters of the $l$-th layer 
in the $k$-th fine-tuned model. 
The goal of model merging is to combine these fine-tuned models, represented by the parameter set $\{\theta_k\}_{k=1}^{K}$, into a single merged model with unified parameters $\theta_\mathrm{merge}$.
A basic merging strategy is weight averaging, expressed as
\begin{equation}
    \theta_\mathrm{merge} = \frac{1}{K} \sum_{k=1}^K \theta_k.
\end{equation} 

Building on this, Task Arithmetic \cite{Ilharco_2022_Editinga} introduces the concept of \textbf{``Task Vector''}. For a task $k$, its task vector $\tv_k$ is defined as $\tv_k = \theta_k - \theta_\text{pre}$. By introducing an additional hyper-parameter $\lambda$, a new model merging paradigm is defined as 

\begin{equation}
    \theta_\mathrm{merge} = \theta_\text{pre} + \lambda \sum_{k=1}^K \tv_k.
    \label{eq:tv}
\end{equation}

Motivated by the task vectors, we further define the concept of \textbf{``Task Feature''}, which captures feature-level deviations caused by the task vector $\tv_t$ relative to the pre-trained model $\theta_\text{pre}$ at different layers. Task feature at layer $l$ when processing an input data $x_k$ from task $k$ is defined as
\begin{equation}
    \tf[x_k]{t}{l} = f^l(x_k; \theta_\text{pre}+\tv_t) - f^l(x_k; \theta_\text{pre}),
    \label{eq:def_tf}
\end{equation}
where function $f^l(\cdot)$ calculates the feature at layer $l$.
And throughout the whole paper, variables with a hat notation (\textit{e.g.}, $\Delta \hat{\theta}^l$ or $\Delta \hat{h}^l(x_t)$) are used to denote the calibrated model.

\begin{figure*}[!t]
    \centering
    \includegraphics[width=1\linewidth]{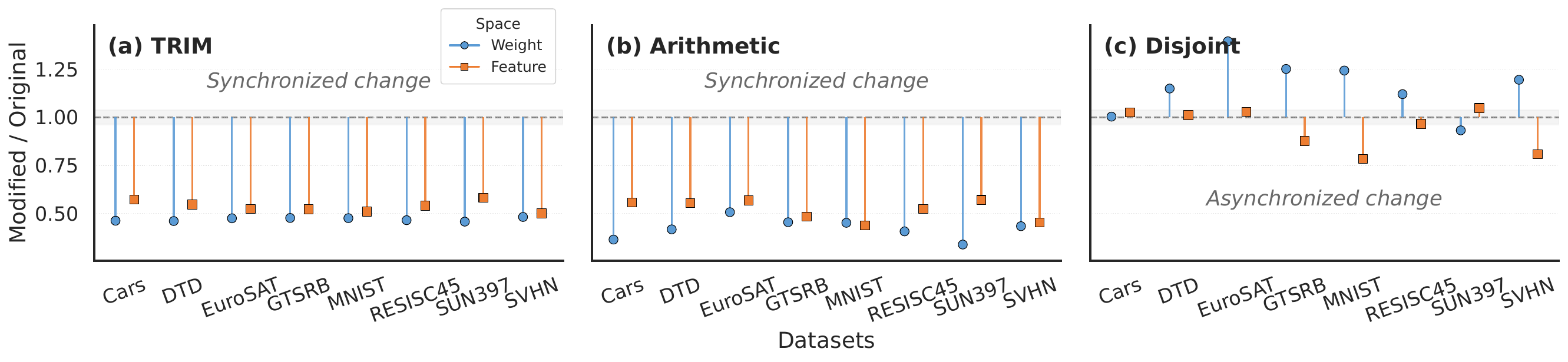}
    \caption{
        Influence of magnitude-altering operations (Left: TRIM, Middle: Arithmetic, Right: Disjoint) on the magnitude of weight and feature spaces. Each point in the figure represents the ratio between the modified magnitude after the operation and the original magnitude before the operation. 
        So, in most cases, variations in feature magnitude closely follow those in weight magnitude, while certain fusion scenarios (c) reveal distinct patterns of variation.
    }
    \label{fig:magnitude-altering op}
\end{figure*}

\subsection{What Causes Magnitude Deviation?} 
\label{sec:unstable_oper}
Magnitude deviation is defined as the difference between the magnitude of the task feature in the merged model and that of the corresponding task feature (\textit{i.e.}, optimal feature) in the specialised model. Magnitude deviation naturally occurs during model merging as a result of operations that modify the norm of model parameters, as well as the intrinsic disentanglement property of task vectors~\cite{Ortiz-Jimenez_2023_Task}. This section provides a comprehensive theoretical analysis of these key factors.

\subsubsection{Magnitude-Altering Operation}
Magnitude-altering operations are defined as those that directly modify the $L_p$-norm of the task vector within the weight space. Among existing merging methods, parameter sparsification and parameter fusion are commonly observed as such operations.

\textbf{Parameter Sparsification.} 
Sparsification methods (\textit{e.g.}, TRIM~\cite{Yadav_2023_TIES-Merging}) prune parameters to reduce interference across models. 
By explicitly setting a portion of the task vector to zero, these operations decrease the overall norm of the task vector, thus shrinking the effective scale of the weight space. 

\textbf{Parameter Fusion.} 
Methods such as Task Arithmetic (TA)~\cite{Ilharco_2022_Editinga} and Disjoint \cite{Yadav_2023_TIES-Merging} fuse task vectors into a single one. 
According to the generalised triangle inequality:
\begin{equation}
    \left\| \mathbb{E} \left [ \tv_k \right ] \right\|_p \leq \mathbb{E} \left [ \left\| \tv_k \right\|_p \right ],
    \label{eq:triangle_generalization}
\end{equation}
where $p \in \mathbb{N}^+$ and $\left\| \mathbb{E} \left [ \tv_k \right ] \right\|_p$ denotes the $L_p$-norm of the averaged task vector, and $\mathbb{E} \left [ \left\| \tv_k \right\|_p \right ]$ represents the expected $L_p$-norm of the task vector prior to merging.  
This inequality demonstrates that any operations similar to weight averaging will inevitably lead to a reduction in their weight space magnitude. Therefore, we can obtain the proposition~\ref{prop:Magnitude-Altering Effect}.

\begin{probox}
    \begin{proposition}[Norm Instability under Fusion]
    \label{prop:Magnitude-Altering Effect}
    Consider a model merging operator $\mathcal{M}(\cdot)$ that performs any form of fusion among the task vectors derived from different models:
    \[
    \tv_\mathrm{merge} = \mathcal{M}( \{\tv_k\}_{k=1}^{K} ).
    \]
    Then, in general,
    \[
    \|\tv_\mathrm{merge}\|_p \neq \mathbb{E}_k[\|\tv_k\|_p], \quad p \in \mathbb{N}^+.
    \]
    In other words, any fusion of task vectors inherently alters their magnitude, leading to potential magnitude variation.
    \end{proposition}
\end{probox}

Under the local linearity assumption, \textbf{the magnitude of task features in each layer is proportional to the magnitude of the corresponding task vector}. Thus, magnitude-altering operations induce analogous magnitude variations in both the weight space and the feature space. 
Visualisation of the effects of magnitude-altering operations (Fig.~\ref{fig:magnitude-altering op}) demonstrates that such operations in the weight space are directly reflected in the feature space.
However, in the right plot of Fig.~\ref{fig:magnitude-altering op}, an increase in overall weight space magnitude is accompanied by a decrease in feature space magnitude, suggesting that additional factors contribute to magnitude deviation.

\subsubsection{Weight Disentanglement}
The abnormal phenomenon in Fig.~\ref{fig:magnitude-altering op} can be explained 
by the \emph{weight disentanglement} property~\cite{Ortiz-Jimenez_2023_Task}, which 
states that the magnitude of a task feature depends not only on the norm of the task vector but also on how well the task vector aligns with the inference data.

This property of \emph{weight disentanglement} can be formalised as follows. Let the $l$-th layer be represented as a differentiable function $f^{l}(\cdot)$ around the pretrained parameter $\theta_{\mathrm{pre}}$. By a first-order expansion to the weight, Eq.~(\ref{eq:def_tf}) can be rewritten as:

\begin{equation}
    \tf[x_k]{t}{l} \approx \mathbf{J}^{l}_{t}(x_k)\,\tv_t^{l},
\end{equation}

where $\mathbf{J}^{l}_{t}(x_k)$ is the Jacobian of $f^{l}$, induced by the input $x_k$, with respect to 
$\Delta \theta_t^{l}$ at $\theta_{\mathrm{pre}}^{\,l}$. 
Taking norms yields,
\begin{equation}
    \|\tf[x_k]{t}{l}\|\ \approx \| \mathbf{J}^{l}_{t}(x_k)\|\,\|\tv_t^{\,l} \| \, \mathrm{cos}(\phi).
\end{equation}

This shows that the magnitude of the task feature depends on two components: 
the norm of the task vector $\tv_t^{\,l}$ and its corresponding Jacobian $\mathbf{J}^{l}_{t}(x_k)$, as well as their alignment, $\mathrm{cos}(\phi)$. Under the Neural Tangent Kernel (NTK)~\cite{NEURIPS2018_5a4be1fa} approximation, the task vector can be written as

\begin{equation}
    \tv_t^{\,l} = \eta\,\mathbb{E}_{x_t\sim p}
    \bigl[
    \mathbf{J}^{l}_{t}(x_t)^\top\,
    g^{l+1}
    \bigr],
\end{equation}
where $p$ is the fine-tuning data distribution and $\eta$ is the learning rate. In the NTK condition, the task vector can be viewed as a one-step update of the gradient on a pretrained model, and by the chain rule, the gradient equals the transpose of the Jacobian multiplied by the upstream gradient, $g^{l+1}$. Substituting this into the feature approximation yields

\begin{equation}
    \tf[x_k]{t}{l}
    \approx 
    \eta\,\mathbf{J}^{l}_{t}(x_k) \,
    \mathbb{E}_{x_t\sim p}
    \bigl[
    \mathbf{J}^{l}_{t}(x_t)^\top
    \bigr]\,
    \mathbb{E}_{x_t\sim p}
    \bigl[
    g^{l+1}
    \bigr].
\end{equation}

When the inference input $x_k$ is sampled from $p$, the Jacobian $\mathbf{J}^{l}_{t}(x_k)$ aligns more strongly with the expected value $\mathbb{E}_{x_t\sim p}[\mathbf{J}^{l}_{t}(x_t)]$, increasing the product $\|\mathbf{J}^{l}_{t}(x_k)\,\tv_t^{\,l}\|$.
In conclusion, \textbf{the task feature magnitude also depends on the alignment between the task vector and the input data}. This explains the observation in Fig.~\ref{fig:magnitude-altering op}, where an increase in weight magnitude coincides with a decrease in feature magnitude, primarily due to a reduction in the aforementioned alignment.

\begin{lembox}
\begin{lemma}[Task Feature Magnitude]
\label{lem:tf_alignment}
Let the $l$-th layer be a differentiable function $f^{l}$ around the pretrained 
parameter $\theta_{\mathrm{pre}}^{\,l}$. For task $t$ and input $x_k$, the corresponding 
task feature at layer $l$ admits the first-order approximation
\[
\tf[x_k]{t}{l} \approx \mathbf{J}^{l}_{t}(x_k)\,\tv_t^{\,l},
\]
where $\mathbf{J}^{l}_{t}(x_k)$ is the Jacobian of $f^{l}$ with respect to 
$\Delta\theta_t^{l}$ evaluated at $\theta_{\mathrm{pre}}^{\,l}$, and $\tv_t^{\,l}$ is the 
task vector at layer $l$.

Under the Neural Tangent Kernel (NTK) approximation, when the distribution of inference data $x_k$ closely matches that of the data $x_t$ utilized for fine-tuning $\tv_t^{\,l}$, the alignment between $\mathbf{J}^{l}_{t}(x_k)$ and $\tv_t^{\,l}$ increases. Consequently, for comparable norms $\|\mathbf{J}^{l}_{t}(x_k)\|$ and $\|\tv_t^{\,l}\|$, this enhanced alignment leads to an increase in $\|\tf[x_k]{t}{l}\|\|$.
\end{lemma}
\end{lembox}

\begin{figure}[!t]
  \centering
   \includegraphics[width=1\linewidth]{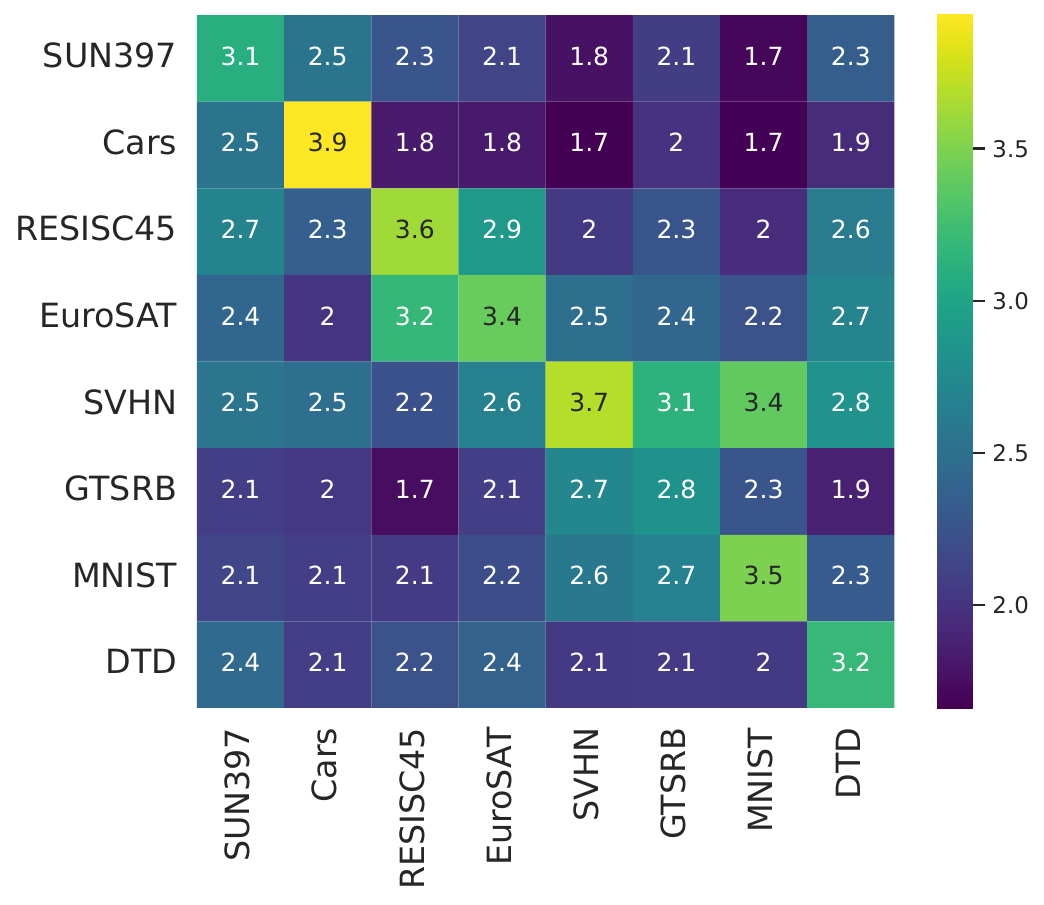}
   \caption{Visualisation of weight disentanglement property: given a merged model, when the task vector of a specific layer $l$ is replaced with that from the specialised model corresponding to the inference task, the feature magnitude of the output at that layer increases. 
   The data in the $i$-th row and $j$-th column represents the increase in the task feature, measured by $L_1$ norm, resulting from substituting $\tv_j^l$ for $\tv_\textrm{merge}^l$ during inference on task $i$.}
   \label{fig:delta_feature_heatmap}
\end{figure}

Such a weight disentanglement phenomenon can be empirically evaluated in Fig.~\ref{fig:delta_feature_heatmap}, supporting that the merged model, who introduce the task-irrelevant task vectors, will induce a lower magnitude of the task feature compared to the specialised model, though the magnitude of their task vector may appear to be the same. Then that's another fundamental source of magnitude deviation observed in merged models.


\begin{figure}[t]
    \centering
    \includegraphics[width=1\linewidth]{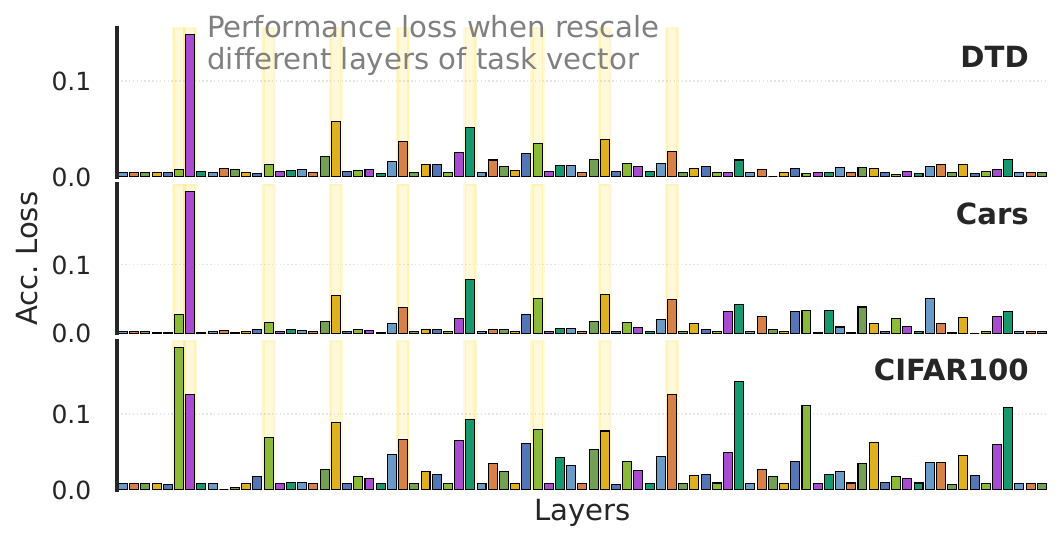}
   \caption{
       Visualisation of the layer sensitivity. Different layers exhibit distinct sensitivity, while the same layers across different task vectors behave similarly. So, a universal task (\eg, CIFAR100) can be utilised to approximate the magnitude-sensitive layers in our target tasks (\eg,  DTD or Cars).
   }
   \label{fig:scale_stability}
\end{figure}

\subsection{How Does Magnitude Deviation Impact Performance?}
\label{sec:Magnitude Influence Performance}
As previously analysed, the merging process inevitably introduces deviation to the magnitude. To evaluate this impact, consider a merged task vector $\tv_{t}$ and a small multiplicative perturbation applied to the task vector of its layer $l$, \textit{i.e.},
\begin{equation}
\tv_{t}^l \to (1+\epsilon) \tv_{t}^l, \quad \text{where } \epsilon > 0 \text{ and } \epsilon \to 0.
\end{equation}

We denote the resulting perturbed model as $\tv_{\epsilon, l}$ and the performance metric as $\mathcal{P}(\cdot)$. 
Then layer sensitivity $s_t^l$ at layer $l$ is defined by evaluating the difference of $\mathcal{P}(\cdot)$ on the task vectors before and after perturbation at the same input $x$. Then $s_t^l$ can be approximated using the first-order Taylor expansion:

\begin{equation}
    s_t^l = \mathcal{P}(\tv_{\epsilon, l}) - \mathcal{P}(\tv_{t}) \approx \mathbf{J}_{\epsilon}^{l}(x)\, \epsilon\tv_{t}^l \label{eq:def_layer_sensitivity}
\end{equation}
where $\mathbf{J}_{\epsilon}^{l}(x)$ is the Jacobian of the performance metric $\mathcal{P}$ with respect to the task vector $\tv_{t}$ at layer~$l$. Therefore, the layer sensitivity of the perturbation to the performance satisfies:

\begin{equation}
   |s_t^l| = | \mathcal{P}(\tv_{\epsilon, l}) - \mathcal{P}(\tv_{t}) | \propto 
   \| \mathbf{J}_{\epsilon}^{l}(x)\, \tv_{t}^l \|.
\end{equation}

By the chain rule, $\mathbf{J}_{\epsilon}^{l}(x)$ at layer $l$ is computed as the product of the Jacobians from layer $l+1$ to layer $L$. 
Since these Jacobian matrices differ across layers, the value $\| \mathbf{J}_{\epsilon}^{l}(x) \tv_{t}^l \|$ also varies, indicating distinct layer sensitivity for each layer.
Thus, \textbf{distinct layers exhibit varying responses to scaling}, suggesting that implementing uniform global scaling across all layers is not an optimal approach.

\begin{thmbox}
    \begin{theorem}[Layer-wise Sensitivity to Task Vector Scaling]
    \label{theorem:layer_sensitivity}
    Give the task vector of a set
    $\tv_k = \{\tv_k^1, \tv_k^2, \dots, \tv_{k}^{L}\}$, 
    where each $\tv_{k}^{l}$ corresponds to the task vector of layer $l$.  
    For a given layer $l$, consider a scaled model defined as
    \[
        \tv_{\epsilon, l} = \{\tv_k^1, \dots, \epsilon \tv_{k}^{l}, \dots, \tv_{k}^{L}\},
    \]
    where $\epsilon > 0$ is a scaling factor. When $\epsilon$ deviates slightly from $1$, the model performance metric $\mathcal{P}(\tv_{\epsilon, l})$ satisfies
    \[
    \mathcal{P}(\tv_{\epsilon, l}) \neq \mathcal{P}(\tv_{\epsilon, l'}) \quad \text{for } l \neq l'.
    \]    
    This result suggests that different layers exhibit varying sensitivities to task vector scaling. 
    \end{theorem}
\end{thmbox}

To empirically validate this, we systematically rescale the magnitude of task vectors from a specialised model at different layers and directly measure the layer sensitivity $s_t^l$, as shown in Fig.~\ref{fig:scale_stability}. 
The results confirm the validity of Theorem~\ref{theorem:layer_sensitivity}.
Besides, \textbf{at the same layer $l$, task vectors associated with different tasks $t$ exhibit strong consistency in $s_t^l$}. 
Specifically, certain common layers are located within sharp regions of the loss landscape across multiple tasks, even when the inference dataset (CIFAR100~\cite{Krizhevsky__Learning}) and the data used for fine-tuning the task vectors are drawn from different distributions.
Small perturbations in the magnitude of these layers, particularly upward perturbations, cause substantial performance degradation. We refer to these layers as \textbf{``Magnitude-Sensitive Layers''}.

\subsection{How to Mitigate Magnitude Deviation?}
\label{sec:Mitigate Magnitude Deviation}
Magnitude deviation reduces the alignment with optimal features, which leads to decreased model performance. Therefore, mitigating magnitude deviation is crucial to achieving stable and generalizable model merging.

Based on \textbf{Lemma~\ref{lem:tf_alignment}}, when a task vector has not been fine-tuned on a distribution similar to that of the inference data (\textit{i.e.}, task-irrelevant weight), the magnitude of the induced task feature remains limited.
From this perspective, the task-irrelevant weights in the merged model contribute minimally to the final task feature. Therefore, the task feature of the merged model can be further decomposed into a task-relevant component with a residual noise term:

\begin{equation}
    \tf[x_k]{\mathrm{merge}}{l} = \eta \tf[x_k]{k}{l} + \epsilon(x_k),  
\end{equation}
where $\tf[x_k]{k}{l}$ represents the task feature of the specialised model, capturing the task-relevant component that truly contributes to performance, and $\epsilon(x_k)$ represents the residual feature contribution from task-irrelevant weights for input $x_k$ from task $k$. The task-specific loss can thus be formulated as:

\begin{align}
    \mathcal{L}_{\mathrm{merge}}(x) 
    &= \| \xi^l \tf[x_k]{\mathrm{merge}}{l} - \tf[x_k]{k}{l} \|^2 \\
    &= \| (\xi^l\eta - 1) \tf[x_k]{k}{l} + \xi^l\epsilon(x) \|^2.
\end{align}

By minimizing \( \mathcal{L}_{\mathrm{merge}}(x) \) with respect to \( \xi^l \), the closed-form solution for \( \xi^l \) can be derived as follows:
\begin{equation}
    \xi^{l,*} = \frac{
    \| \Delta h_k^{\,l}(x_k) \|^2 \eta - \langle \Delta h_k^{\,l}(x_k), \epsilon(x) \rangle
    }{
    \| \Delta h_k^{\,l}(x_k) \|^2 \eta^2 + \| \epsilon(x) \|^2
    }.
\end{equation}

In model merging, current methods progressively suppress the residual interference term $\epsilon(x)$, thereby enhancing the weight disentanglement. In the idealized scenario where $\epsilon(x) \to 0$, the optimal scaling coefficient simplifies to:

\begin{equation}
    \xi^l = \frac{1}{\eta} = \frac{\| \tf[x_k]{k}{l} \|}{\| \tf[x_k]{\mathrm{merge}}{l} \|} \quad \text{when} \ \epsilon(x) \to 0.
\end{equation} 

This indicates that when the magnitude of the merged model's task feature most closely approximates that of the task-specialised model, their respective task-specific capabilities also converge to the greatest extent.
We formally summarise this finding as the following theorem:

\begin{thmbox}
    \begin{theorem}[Optimal Layer-wise Scaling]
    \label{prop:optimal_feature_norm}
    Assume the model fully satisfies the \textit{weight disentanglement} property (\textit{i.e.}, residual term $\epsilon(x) \to 0$).
    Then, for any layer $l$, the merged model achieves its optimal task performance when the magnitude of its task feature matches that of the corresponding specialised model with the optimal layer-wise scaling coefficient $\xi^l$ in the feature space:
    \[
        \xi^l = \frac{\| \tf[x_k]{k}{l} \|}{\| \tf[x_k]{\mathrm{merge}}{l} \|}
    \]
    Therefore, minimising the feature-norm discrepancy (\textit{i.e.}, magnitude deviation) between the merged and specialised models effectively improves task-specific performance.
    \end{theorem}
\end{thmbox}

\section{Methodology}
\label{sec:method}
\textbf{Theorem~\ref{theorem:layer_sensitivity}} reveals that sensitivity to magnitude perturbations varies by layer. 
Thus, scale calibration should be performed per layer rather than through uniform global rescaling. Specifically, it is important to first identify the layers most sensitive to magnitude changes (\textit{i.e.}, magnitude-sensitivity layers) and suppress their magnitudes. Applying the same scaling to all layers can significantly degrade performance.


Based on this, a layer-wise approach to scale calibration is introduced to ensure that the merged model best preserves the optimal feature magnitude established in \textbf{Theorem~\ref{prop:optimal_feature_norm}}.
Specifically, we propose a simple yet effective calibration framework, MAGIC: \(\hat{\delta}^l = \xi^l \delta^l\), where \( \delta^l\) can be represented as task feature $\tf[x_t]{}{l}$ or task vector $\tv_{k}^{l}$. This framework applies a calibration coefficient \(\xi^l\) across layers, yielding three effective variants: WSC, FSC and DSC.

\subsection{Magnitude-Sensitivity Layers Selection}
As discussed in Sec.~\ref{sec:Magnitude Influence Performance}, certain magnitude-sensitive layers tend to reside within steep regions of the loss landscape, and this layer-wise sensitivity remains consistent when performing inference on multiple distinct tasks. 

To identify such layers in merged models, a general dataset (such as CIFAR100 or ImageNet) is utilised, which may originate from a data distribution different from that used for fine-tuning the task vector.
And then compute the layer sensitivity $s_\mathrm{avg}^l$ of the merged model through weight averaging based on its definition in Eq.~(\ref{eq:def_layer_sensitivity}). 
Layers are then ranked by their layer sensitivity, with the top-$\alpha$ layers (those most sensitive to magnitude amplification) grouped into set $\mathcal{A}$, referred to as magnitude-sensitive layers.
\begin{equation}
    \mathcal{A}=\left\{l \mid \left|\left\{k \mid s_\mathrm{avg}^k<s_\mathrm{avg}^l\right\} \right| <\alpha\right\}.
    \label{eq:add_to_set_A}
\end{equation}

\subsection{Feature Space Calibration (FSC)}
\label{sec:feature_cal}
As demonstrated in \textbf{Theorem~\ref{prop:optimal_feature_norm}}, under the assumption of the full weight disentanglement, the performance of the merged model closely approximates that of the specialised model if and only if their task features exhibit identical magnitudes. 
In multi-task scenarios, optimal average performance is achieved by matching the task feature magnitudes in the merged model to the average magnitude of those in the specialised models.
Thus, the calibration coefficient \(\xi^l\) can be calculated as:

\begin{equation}
    \xi^l = \frac{1}{K} \sum_{k=1}^K \frac{\| \tf[x_k]{k}{l} \|}{\| \tf[x_k]{\mathrm{merge}}{l} \|}.
    \label{eq:FSC}
\end{equation}

Since  Sec.~\ref{sec:Magnitude Influence Performance} demonstrates that certain layers are sensitive to magnitude amplification, directly multiplying the calibration coefficient by the task feature of each layer may pose potential risks.
To mitigate this risk, a conservative calibration strategy is employed. Specifically, for layers in the magnitude-sensitive set $\mathcal{A}$, calibrations are applied only when $\xi^l < 1$ to prevent performance degradation caused by increasing the magnitude of these sensitive layers. For layers not included in set $\mathcal{A}$, calibrations are performed only when $\xi^l > 1$, thereby maintaining adequate performance. This procedure can be formally represented by the XOR operation \(\oplus\) as follows:

\begin{equation}
    \hat{\tf[x_t]{}{l}} = 
    \begin{cases} 
    \xi^l \tf[x_t]{}{l}, & \text{if } (\xi^l > 1) \oplus (l \in \mathcal{A}) \\
    \tf[x_t]{}{l}, & \text{otherwise}
    \end{cases}.
    \label{eq:WSC}
\end{equation}

\begin{algorithm}[t]
    \caption{Feature Space Calibration (FSC)}\label{alg:FSC}
    \begin{algorithmic}[1]
        \STATE \textbf{Input:} Merged weights $\theta_{\mathrm{merge}}$; task feature $\tf[x_k]{\mathrm{merge}}{l}$, $\{ \tf[x_k]{k}{l}\}_{k=1}^K$; Magnitude-sensitive layer set $\mathcal{A}$.
        \STATE \textbf{Output:} Calibrated merged model $\hat{\theta}_{\mathrm{merge}}$.
        \FOR{$l = 1, \dots, L$}
            \STATE $\xi_f^l \gets \frac{1}{K} \sum_{k=1}^K \frac{\| \tf[x_k]{k}{l} \|}{\| \tf[x_k]{\mathrm{merge}}{l} \|}.$ (Eq.~\ref{eq:FSC})
            \IF{not $(\xi_f^l > 1) \oplus (l \in \mathcal{A})$}
                 \STATE $\xi_f^l \gets 1$
            \ENDIF
        \ENDFOR
        
        \Statex \textit{\textcolor{StepA}{Integrate $\xi_f^l$ into the model's forward function.}}
        \STATE $\hat{\theta}_\mathrm{merge} \gets \textsc{Integrate}\big(\hat{\theta}_\mathrm{merge}, \{\xi_f^l\}_{l=1}^L\big)$
        \STATE {\bfseries Return:} $\hat{\theta}_\mathrm{merge}$
    \end{algorithmic}
\end{algorithm}

\subsection{Weight Space Calibration (WSC)}
\label{sec:weight_cal}
Feature-space calibration effectively addresses magnitude deviations; however, it relies on additional input data to extract features. This dependency raises a fundamental question:  \textit{Is it possible to perform calibration directly in the weight space, thereby eliminating the need for additional data?}

Inspired by the Spherical Linear Interpolation (SLERP) formulation~\cite{Shoemake_1985_Animating}, we extend this idea to a multi-model merging context through the geometric interpretation of a \textit{hyperellipsoid} centred at the origin. 
Given that task vectors between different models are approximately orthogonal~\cite{Ilharco_2022_Editinga, tang2023parameter} in high-dimensional space, these vectors can serve as the principal axes of a hyperellipsoid that characterises the layer-wise weight space. The objective is to ensure that the task vector of the merged model lies on the corresponding hyperellipsoid, thereby constraining the magnitude appropriately.

Specifically, for each layer $l$, the merged task vector $\tv_{k}^{l}$ can be decomposed into two orthogonal components. 
The projection of $\tv_{k}^{l}$ onto each known task axis $\tv_k^l$ is:
\begin{equation}
    \operatorname{Proj}_{\tv_k^l}\left(\tv_{k}^{l}\right)
    = \frac{\left\langle \tv_{k}^{l}, \tv_k^l \right\rangle}
    {\left\|\tv_k^l\right\|^2} \tv_k^l.    
\end{equation}

The residual component $\tv_k^{l,\perp}$ that is orthogonal to all task directions is then given by:
\begin{equation}
    \tv_k^{l,\perp} = \tv_{k}^{l} - \sum_{k=1}^{K} 
\operatorname{Proj}_{\tv_k^l}\left(\tv_{k}^{l}\right).
\end{equation}

The construction of the hyperellipsoid within this space imposes the following constraint:
\begin{equation}
    \sum_{k=1}^K \frac{ \left\| \operatorname{proj}_{\tv_k^l} \left(\tv_{k}^{l} \right) \right\|^2 }{ \left\| \tv_k^l \right\|^2} + \frac{\left\| \tv_k^{l,\perp} \right\|^2}{\bar{M}^2}= \frac{1}{\left(\xi^l\right)^2},
    \label{eq:WSC}
\end{equation}
where $\bar{M} = \frac{1}{K}\sum_{k=1}^K \|\tv_k^l\|$ denotes the average task-vector magnitude within the layer.  
Here, the left-hand term defines a hyperellipsoid spanned by the known task directions, 
and the right-hand term represents an isotropic hypersphere in the residual subspace. 
By solving this constraint, we obtain the layer-specific weight calibration coefficient $\xi^l$, and then apply the conservative calibration strategy analogous to Eq.~(\ref{eq:WSC}).

\begin{algorithm}[t]
    \caption{Weight Space Calibration (WSC)}\label{alg:WSC}
    \begin{algorithmic}[1]
        \STATE \textbf{Input:} Pre-trained weights $\theta_{\mathrm{pre}}$; layer-wise task vector $\tv_{\mathrm{merge}}^{l}$, $\{ \tv_k^{\,l} \}_{k=1}^K$; Magnitude-sensitive layer set $\mathcal{A}$.
        \STATE \textbf{Output:} Calibrated merged model $\hat{\theta}_{\mathrm{merge}}$.
        
        \FOR{$l = 1, \dots, L$}
            \STATE $\xi_w^l \gets \textsc{ScaleToHyperEllipsoid}\big($\\
            \hspace{0.2cm} $\tv_{\mathrm{merge}}^{l}, \{\tv_k^{\,l}\}_{k=1}^K\big)$ (Eq.~\ref{eq:WSC})

            \IF{not $(\xi_w^l > 1) \oplus (l \in \mathcal{A})$}
                 \STATE $\xi_w^l \gets 1$
            \ENDIF
            
            \STATE $\Delta \hat{\theta}_{\mathrm{merge}}^l \gets \xi_w^l \,\tv_{\mathrm{merge}}^{l}$
        \ENDFOR

        \Statex \textit{\textcolor{StepA}{Integrate merged task vector with pre-trained weight.}}
        \STATE $\hat{\theta}_\mathrm{merge} \gets \textsc{Recompose}\big(\{\Delta \hat{\theta}_{\mathrm{merge}}^l\}_{l=1}^L, \theta_{\mathrm{pre}}\big)$ (Eq.~\ref{eq:tv})
        \STATE {\bfseries Return:} $\hat{\theta}_\mathrm{merge}$
    \end{algorithmic}
\end{algorithm}

\begin{algorithm}[t]
    \caption{Dual-Space Calibration (DSC)}\label{alg:DSC}
    \begin{algorithmic}[1]
        \STATE \textbf{Input:} 
        Pre-trained weights $\theta_{\mathrm{pre}}$;
        Merged weights $\theta_{\mathrm{merge}}$; 
        layer-wise task vector $\tv_{\mathrm{merge}}^{l}$, $\{ \tv_k^{\,l} \}_{k=1}^K$ and task feature $\tf[x_k]{\mathrm{merge}}{l}$, $\{ \tf[x_k]{k}{l}\}_{k=1}^K$.
        \STATE \textbf{Output:} Calibrated merged model $\hat{\theta}_{\mathrm{merge}}$.
        
        \STATE $\mathcal{A} \gets$ Computed by Eqs.~(\ref{eq:def_layer_sensitivity}) and (\ref{eq:add_to_set_A}).

        \STATE $\!\hat{\theta}_{\mathrm{merge}} \! \gets \! \textsc{WSC}(\theta_{\mathrm{pre}},\, \tv_{\mathrm{merge}}^{l},\, \{\tv_k^{\,l}\},\, \mathcal{A})$ (Alg.~\ref{alg:WSC})\\
        \STATE $\!\hat{\theta}_{\mathrm{merge}} \! \gets \! \textsc{FSC}(\hat{\theta}_{\mathrm{merge}}, \! \tf[x_k]{\mathrm{merge}}{l}, \! \{\!\tf[x_k]{k}{l}\!\}\!,\! \mathcal{A})\!$ (Alg.~\ref{alg:FSC})\\
        
        
        \STATE {\bfseries Return:} $\hat{\theta}_\mathrm{merge}$
    \end{algorithmic}
\end{algorithm}

\subsection{Dual Space Calibration (DSC)}
Given the distinct advantages offered by calibrations in both weight and feature spaces, integrating these approaches enables effective dual space calibration. Specifically, DSC utilizes the calibration formulation \(\hat{\delta}^l = \xi^l \delta^l\) in two stages: first, by setting \(\delta^l = \tv_{k}^{l}\) to perform calibration within the weight space, and subsequently by setting \(\delta^l = \tf[x_t]{}{l}\) to conduct calibration in the feature space. This dual application facilitates precise estimation of the calibration coefficients. The step-by-step procedure for DSC is outlined in Alg.~\ref{alg:DSC}.

In summary, our method enhances the stability of merged models by applying calibration coefficients to different layers across various spaces. Because our approach does not involve complex computations such as gradient descent and requires only a small amount of data to estimate magnitude variation in feature space, its minimal computational cost and data requirements make it well-suited for real-world applications.


\begin{table*}[t]
\centering
\caption{Multi-task performance (\%) when merging ViT-B/32 models on eight tasks, where ``specialised'' refers to individual task-specific models, and ``weight averaging'' indicates the simple averaging of model parameters. }
\footnotesize
\begin{tabularx}{\linewidth}{l|l*{8}{C}c}
\toprule
& Method           & SUN397 & Cars & RESISC45 & EuroSAT & SVHN & GTSRB & MNIST & DTD  & Avg Acc              \\ \midrule
& Pretrained       & 62.3   & 59.7 & 60.7     & 45.5    & 31.4 & 32.6  & 48.5  & 43.8 & 48.0                 \\
& Individual       & 75.3   & 77.7 & 96.1     & 99.7    & 97.5 & 98.7  & 99.7  & 79.4 & 90.5                 \\
& Traditional MTL~\cite{Yang_2024_AdaMerging}  & 73.9   & 74.4 & 93.9     & 98.2    & 95.8 & 98.9  & 99.5  & 77.9 & 88.9                 \\ \midrule
\multirow{6}{*}{\rotatebox{90}{Baseline}} & Weight Averaging & 65.3   & 63.4 & 71.4     & 71.7    & 64.2 & 52.8  & 87.5  & 50.1 & 65.8                 \\
& TA$_\text{ ICLR'23}$~\cite{Ilharco_2022_Editinga}               & 55.2   & 54.9 & 66.7     & 78.9    & 80.2 & 69.7  & 97.3  & 50.4 & 69.1                 \\
& TIES$_\text{ NeurIPS'23}$~\cite{Yadav_2023_TIES-Merging}             & 59.8   & 58.6 & 70.7     & 79.7    & 86.2 & 72.1  & 98.3  & 54.2 & 72.4                 \\
& TSV-M$_\text{ CVPR'25}$~\cite{Gargiulo_2024_Task}            & 69.1   & 70.8 & 85.5     & 94.2    & 92.0   & 91.9  & 99.2  & 69.1 & 84.0                   \\
& Iso-C$_\text{ ICML'25}$~\cite{Marczak_2025_No}            & 74.8   & 74.1 & 87.9     & 92.9    & 83.1 & 86.0    & 98.2  & 67.9 & 83.1                 \\
& Iso-CTS$_\text{ ICML'25}$~\cite{Marczak_2025_No}          & 74.4   & 74.4 & 87.2     & 90.4    & 76.8 & 83.3  & 97.4  & 67.0   & 81.3                 \\ \midrule
\multirow{5}{*}{\rotatebox{90}{Ours}} & TA w/ DSC        & 66.8   & 67.0   & 79.7     & 92.0      & 86.4 & 81.1  & 98.6  & 59.7 & 78.9$_\text{ (9.8↑)}$        \\
& TIES w/ DSC      & 67.0     & 66.5 & 79.7     & 88.1    & 89.0   & 76.1  & 98.7  & 59.5 & 78.1$_\text{ (5.7↑)}$        \\
& TSV-M w/ DSC     & 72.1   & 75.0   & 87.7     & 95.6    & 89.9 & 91.1  & 99.1  & 70.2 & 85.1$_\text{ (1.1↑)}$         \\
& Iso-C w/ DSC     & 74.9   & 74.2 & 88.7     & 96.4    & 87.8 & 90.9  & 99.1  & 70.8 & {\uline{85.3$_\text{ (2.2↑)}$}}    \\
& Iso-CTS w/ DSC   & 77.6   & 77.3 & 91.6     & 95.0      & 80.0   & 89.9  & 98.7  & 74.4 & \textbf{85.6$_\text{ (4.3↑)}$} \\ \bottomrule
\end{tabularx}
\label{tab:main_tab}
\end{table*}

\begin{table*}[t]
\centering
\caption{Weight and feature space calibrated multi-task performance (\%) when merging ViT-B/32 models on eight tasks.}
\footnotesize
\begin{tabularx}{\linewidth}{l *{8}{C} c}
    \toprule
    Method         & SUN397 & Cars & RESISC45 & EuroSAT & SVHN & GTSRB & MNIST & DTD  & Avg Acc              \\ \midrule
    TSV-M w/ WSC   & 71.9   & 74.8 & 87.9     & 95.6    & 90.5 & 91.4  & 99.2  & 71.0   & \textbf{85.3$_\text{ (1.3↑)}$} \\
    Iso-C w/ WSC   & 75.2   & 74.8 & 88.3     & 92.6    & 78.7 & 83.0  & 97.6  & 68.2 & {\uline{82.3$_\text{ (0.8↓)}$}}    \\
    Iso-CTS w/ WSC & 74.9   & 75.0 & 87.1     & 89.9    & 71.6 & 80.2  & 96.4  & 66.6 & 80.2$_\text{ (1.1↓)}$          \\ \midrule
    TSV-M w/ FSC   & 70.2   & 71.8 & 86.2     & 94.7    & 91.3 & 91.8  & 99.3  & 69.4 & 84.3$_\text{ (0.3↑)}$           \\
    Iso-C w/ FSC   & 75.6   & 74.2 & 89.8     & 96.8    & 88.9 & 92.4  & 99.2  & 73.6 & \textbf{86.3$_\text{ (3.2↑)}$} \\
    Iso-CTS w/ FSC & 77.2   & 76.2 & 91.2     & 95.8    & 83.9 & 91.6  & 98.9  & 74.4 & {\uline{86.2$_\text{ (4.9↑)}$}}    \\ \bottomrule
\end{tabularx}
\label{tab:main_tab2}
\end{table*}

\begin{table*}[t]
\centering
\caption{Multi-task performance (\%) when merging models with different backbones on 8 tasks.}
\footnotesize
\begin{tabularx}{\linewidth}{ll *{8}{C} c}
\toprule
Backbone                   & Method                                 & SUN397                       & Cars                         & RESISC45                     & EuroSAT                      & SVHN                         & GTSRB                        & MNIST                        & DTD                          & Avg Acc                            \\ \midrule
& Iso-C                                  & 78.1                         & 82.3                         & 91.9                         & 96.9                         & 88.3                         & 91.9                         & 98.8                         & 71.9                         & 87.5                               \\
& Iso-CTS                                & 77.9                         & 83.2                         & 92.0                         & 96.4                         & 84.9                         & 91.3                         & 98.4                         & 71.1                         & 86.9                               \\
& \cellcolor[HTML]{EFEFEF}Iso-C w/ DSC$_\text{ (Ours)}$   & \cellcolor[HTML]{EFEFEF}77.6 & \cellcolor[HTML]{EFEFEF}82.2 & \cellcolor[HTML]{EFEFEF}92.5 & \cellcolor[HTML]{EFEFEF}97.9 & \cellcolor[HTML]{EFEFEF}92.2 & \cellcolor[HTML]{EFEFEF}94.5 & \cellcolor[HTML]{EFEFEF}99.3 & \cellcolor[HTML]{EFEFEF}77.2 & \cellcolor[HTML]{EFEFEF}\uline{$89.2_\text{ (1.7↑)}$} \\
\multirow{-4}{*}{ViT-B/16} & \cellcolor[HTML]{EFEFEF}Iso-CTS w/ DSC$_\text{ (Ours)}$ & \cellcolor[HTML]{EFEFEF}79.1 & \cellcolor[HTML]{EFEFEF}84.6 & \cellcolor[HTML]{EFEFEF}94   & \cellcolor[HTML]{EFEFEF}98.1 & \cellcolor[HTML]{EFEFEF}90.1 & \cellcolor[HTML]{EFEFEF}95.8 & \cellcolor[HTML]{EFEFEF}99.1 & \cellcolor[HTML]{EFEFEF}78.9 & \cellcolor[HTML]{EFEFEF}\textbf{90.0$_\text{ (3.1↑)}$} \\ \midrule
& Iso-C                                  & 82.0                         & 90.9                         & 94.8                         & 98.7                         & 91.4                         & 95.5                         & 99.3                         & 79.1                         & 91.4                               \\
& Iso-CTS                                & 81.3                         & 91.1                         & 94.7                         & 98.6                         & 89.4                         & 95.3                         & 99.2                         & 77.9                         & 90.9                               \\
& \cellcolor[HTML]{EFEFEF}Iso-C w/ DSC$_\text{ (Ours)}$   & \cellcolor[HTML]{EFEFEF}82.2 & \cellcolor[HTML]{EFEFEF}89.7 & \cellcolor[HTML]{EFEFEF}94.6 & \cellcolor[HTML]{EFEFEF}98.9 & \cellcolor[HTML]{EFEFEF}95.3 & \cellcolor[HTML]{EFEFEF}98.0 & \cellcolor[HTML]{EFEFEF}99.5 & \cellcolor[HTML]{EFEFEF}81.0 & \cellcolor[HTML]{EFEFEF}\uline{$92.4_\text{ (1.0↑)}$} \\
\multirow{-4}{*}{ViT-L/14} & \cellcolor[HTML]{EFEFEF}Iso-CTS w/ DSC$_\text{ (Ours)}$ & \cellcolor[HTML]{EFEFEF}83.3 & \cellcolor[HTML]{EFEFEF}91.3 & \cellcolor[HTML]{EFEFEF}95.9 & \cellcolor[HTML]{EFEFEF}99.0 & \cellcolor[HTML]{EFEFEF}95.5 & \cellcolor[HTML]{EFEFEF}98.5 & \cellcolor[HTML]{EFEFEF}99.5 & \cellcolor[HTML]{EFEFEF}83.1 & \cellcolor[HTML]{EFEFEF}\textbf{93.3$_\text{ (2.4↑)}$} \\ \bottomrule
\end{tabularx}
\label{tab:backbone_tab}
\end{table*}

\section{Experiments}
\label{sec:exp}

\subsection{Setting and Implementation}
We evaluate our multi-task vision and language models by calculating the average accuracy across the test sets of all tasks. Details are provided in the following.

\textbf{Baseline and Datasets}.
Our proposed method is mainly compared with other training-free merging methods, including Weight Averaging (WA), Task Arithmetic (TA)~\cite{Ilharco_2022_Editinga}, Ties-Merging (TIES)~\cite{Yadav_2023_TIES-Merging}, TSV-M~\cite{Gargiulo_2025_Task} and Iso-C~\cite{Marczak_2025_No}. Furthermore, its superiority is also demonstrated over Layer-wise AdaMerging (LWA) \cite{Yang_2024_AdaMerging}, a method that requires training.

For computer vision (CV) tasks, following previous works \cite{Ilharco_2022_Editinga, Yang_2024_AdaMerging}, we test on eight primary image classification datasets, utilising CLIP~\cite{radford2021learning} as the pretrained backbone: SUN397 \cite{Xiao_2016_SUNa}, Cars \cite{Krause_2013_3Da}, RESISC45 \cite{Cheng_2017_Remote}, EuroSAT \cite{Helber_2019_EuroSATa}, SVHN \cite{Netzer__Reading}, GTSRB \cite{Stallkamp_2011_German}, MNIST \cite{MNIST} and DTD \cite{Cimpoi_2014_Describinga}. 

For natural language processing (NLP) tasks, BERT~\cite{devlin2019bert} is fine-tuned on four binary classification datasets: AG News (AN)~\cite{Zhang_2015_Character-level}, Rotten Tomatoes (RT)~\cite{Pang_2005_Seeing}, CoLA~\cite{Warstadt_2019_Neural}, and SMS~\cite{Almeida_2011_Contributions}. Subsequently, parameter-efficient fine-tuning is applied to T0 on the datasets RTE~\cite{dagan2005pascal}, CB~\cite{de2019commitmentbank}, Winogrande (Winog.)~\cite{sakaguchi2020winogrande}, WiC~\cite{pilehvar2018wic}, WSC~\cite{levesque2012winograd}, COPA~\cite{roemmele2011choice}, H-SWAG~\cite{zellers2019hellaswag}, StoryCloze~\cite{sharma2018tackling}, and ANLI~\cite{nie2019adversarial}.  
Finally, to assess the performance of merged models on large language models (LLMs) such as Llama, the fine-tuned models WizardMath-7B-V1.0~\cite{luo2023wizardmath} and Llama-2-7b-chat-hf~\cite{touvron2023llama} are merged and evaluated on two benchmarks: AlpacaEval~\cite{alpaca_eval} and GSM8K~\cite{cobbe2021gsm8k}.

\textbf{Implementation Details.} 
Following prior work \cite{Yang_2024_AdaMerging},  ViT-B/32 CLIP serves as the default visual encoder, and the scaling ratio \(\lambda\) is set to 0.3 for both Task Arithmetic and Ties-Merging.  
Ties-Merging retains the top 20\% of parameters, consistent with the original implementation.
Our introduced hyperparameter \(\alpha\), which determines the size of the magnitude-sensitive layer set \(\mathcal{A}\), is fixed at 10 in all experiments unless otherwise specified, and the perturbation parameter is set to \(\epsilon = 1.1\).
All experiments are conducted on a single NVIDIA A800 GPU.
Additionally, for CV experiments, we follow the checkpoints released in Task Arithmetic \cite{Ilharco_2022_Editinga}. \uline{Some merging baselines such as Iso-C rely on alternative checkpoints that were retrained independently, which may lead to differences between our reproduced results and those reported in their original papers.}

\begin{table*}[thb]
    \centering
    \caption{Performance comparison of merging T0 ($\mathrm{IA}^3$ PEFT) models.}
    \footnotesize
    \begin{tabularx}{\linewidth}{l *{11}{C} c}
    \toprule
    Method    & rte            & cb             & winog.     & wic            & wsc            & copa           & h-swag         & storycloze    & anli-r1        & anli-r2        & anli-r3        & Avg. Acc.      \\ \midrule
    TA        & 71.9          & 56.2          & 53.1          & 29.6          & 65.6          & 78.1          & 46.9          & 87.5          & 46.9          & 28.1          & 25.0          & 53.5          \\
    TIES      & 71.9          & 59.4          & 53.1          & 31.2          & 62.5          & 79.6          & 46.9          & 87.5          & 46.9          & 31.2          & 25.0          & 54.1          \\
    DARE      & 71.9          & 59.4          & 53.1          & 29.6          & 62.5          & 78.1          & 43.8          & 87.5          & 46.9          & 28.1          & 25.0          & 53.3          \\
    TSV-M      & —              & —              & —              & —              & —              & —              & —              & —              & —              & —              & —              & —              \\
    Iso-C     & —              & —              & —              & —              & —              & —              & —              & —              & —              & —              & —              & —              \\
    Iso-CTS   & —              & —              & —              & —              & —              & —              & —              & —              & —              & —              & —              & —              \\
    \rowcolor[HTML]{EFEFEF} 
    TA w/ WSC$_\text{ (Ours)}$ & 71.9          & 59.4          & 53.1          & 29.6          & 62.5          & 79.6          & 46.9          & 87.5          & 50.0          & 31.2          & 25.0          & \textbf{54.2$_\text{ (0.7↑)}$}          \\
    \rowcolor[HTML]{EFEFEF} 
    TIES w/ WSC$_\text{ (Ours)}$ & 71.9          & 59.4          & 53.1          & 29.6          & 65.6          & 78.1          & 46.9          & 87.5          & 46.9          & 31.2          & 25.0          & \uline{54.1$_\text{ (0.0↑)}$}          \\
    \rowcolor[HTML]{EFEFEF} 
    DARE w/ WSC$_\text{ (Ours)}$ & 71.9          & 59.4          & 53.1          & 29.6          & 64.1          & 78.1          & 46.9          & 87.5          & 46.9          & 31.2          & 25.0          & 54.0$_\text{ (0.7↑)}$          \\
    \bottomrule
    \end{tabularx}
    \label{tab:T0}
\end{table*}

\begin{table}[!t]
    \centering
    \caption{Performance comparison of merging BERT models.}
    \footnotesize
    \begin{tabularx}{\linewidth}{@{}lCCCCc@{}}
    \toprule
        Method    & AN & RT & Cola & SMS & Avg. Acc.  \\ \midrule
        TA        & 51.5          & 51.9           & 68.0          & 14.5          & 46.5     \\
        TIES      & 46.5          & 74.6           & 55.4          & 87.6          & 66.0     \\
        DARE      & 47.5          & 55.4           & 69.2          & 44.1          & 54.0     \\
        TSV-M     & 48.6          & 61.6           & 72.7          & 88.4          & 67.8     \\
        Iso-C     & 43.5          & 48.0           & 62.9          & 54.7          & 52.3     \\
        Iso-CTS   & 44.6          & 49.0           & 65.1          & 46.6          & 51.3     \\
        \rowcolor[HTML]{EFEFEF} 
        TA w/ DSC$_\text{ (Ours)}$ & 46.9 & 67.3 & 59.8 & 97.7 & \uline{67.9$_\text{ (21.4↑)}$} \\ 
        \rowcolor[HTML]{EFEFEF} 
        TIES w/ DSC$_\text{ (Ours)}$ & 47.6 & 75.0 & 53.5 & 92.7 & 67.2$_{\,\,\text{ (1.2↑)}}$ \\ 
        \rowcolor[HTML]{EFEFEF} 
        DARE w/ DSC$_\text{ (Ours)}$ & 47.3 & 70.2 & 53.3 & 97.8 & 67.2$_\text{ (13.2↑)}$ \\ 
        \rowcolor[HTML]{EFEFEF} 
        TSV-M w/ DSC$_\text{ (Ours)}$ & 46.7 & 65.4 & 67.5 & 97.1 & \textbf{69.2$_{\,\,\text{ (1.4↑)}}$} \\
        \rowcolor[HTML]{EFEFEF} 
        Iso-C w/ DSC$_\text{ (Ours)}$ & 50.4 & 59.6 & 59.6 & 96.4 & 66.5$_\text{ (14.2↑)}$ \\ 
        \rowcolor[HTML]{EFEFEF} 
        Iso-CTS w/ DSC$_\text{ (Ours)}$ & 49.1 & 55.4 & 62.5 & 88.2 & 63.8$_\text{ (12.5↑)}$ \\ 
        \bottomrule
    \end{tabularx}
    \label{tab:BERT}
\end{table}

\subsection{Calibration Effectiveness}
The effectiveness of the proposed magnitude calibration is validated in both computer vision (CV) and natural language processing (NLP) tasks.  
\textbf{For CV tasks:} 
Tab.~\ref{tab:main_tab} and~\ref{tab:main_tab2} report the performance of merged models on eight datasets, demonstrating that the calibration method significantly enhances previous SOTA approaches. Furthermore, the proposed calibration remains robust across various backbone architectures, as evidenced in Tab.~\ref{tab:backbone_tab}.  
\textbf{For NLP tasks:} 
We first evaluate our calibration method on small-scale models, such as BERT (Tab.~\ref{tab:BERT}), showing consistent performance improvements over previous approaches. To further assess effectiveness, we conduct additional experiments on large-scale models like Llama (Tab.~\ref{tab:Llama}) and IA$^3$ PEFT fine-tuning scenarios (Tab.~\ref{tab:T0}) using WSC as a case study. The results demonstrate our robustness. 
Notably, because IA$^3$ produces vector-form merging parameters, previous SVD-based methods (\textit{i.e.}, TSV-M, Iso-C, Iso-CTS) cannot be applied.
In most settings, our calibration method achieves and surpasses SOTA performance.

\begin{table}[!t]
    \caption{Performance comparison of merging Llama-2-7B models.}
    \centering
    \footnotesize
    \begin{tabularx}{\linewidth}{lCCc}
    \toprule
    Method           & AlphcaEval & GSM8K & Avg.           \\ \midrule
    TA               & 42.6                 & 43.8          & 43.2          \\
    TIES             & 40.3                 & 45.0          & 42.7          \\
    DARE             & 37.2                 & 47.0          & 42.1          \\
    TSV-M            & 48.1                 & 49.7          & 48.9          \\
    Iso-C            & 29.5                 & 42.0          & 35.7          \\
    Iso-CTS          & 24.0                 & 38.7          & 31.4          \\
    \rowcolor[HTML]{EFEFEF} 
    TA w/ WSC$_\text{ (Ours)}$ & 51.2        & 51.1 & \textbf{51.2$_\text{ (8.0↑)}$} \\ \bottomrule
    \end{tabularx}
    \label{tab:Llama}
\end{table}

\subsection{Further Analysis}

\begin{figure}[htb]
  \centering
   \includegraphics[width=1\linewidth]{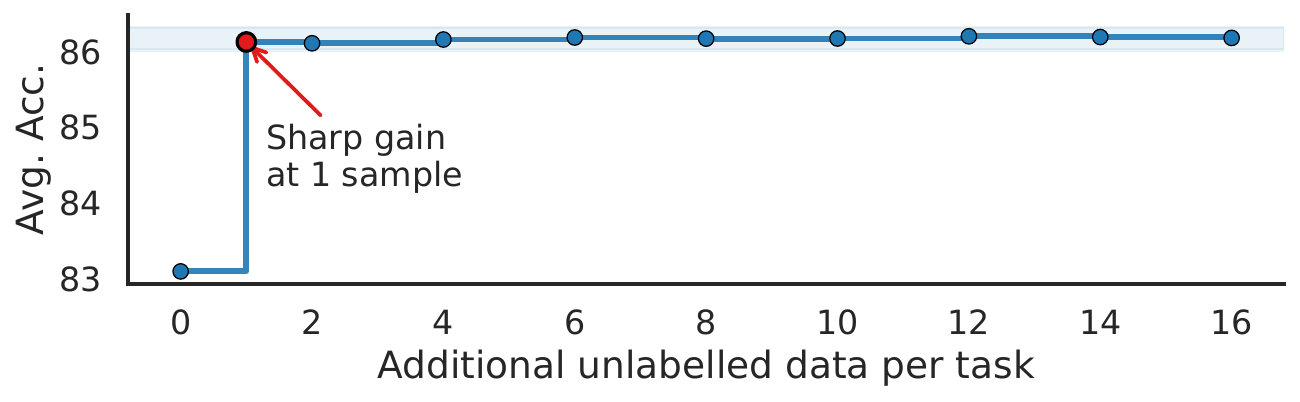}
   \caption{Impact of unlabeled data quantity on Iso-C w/ DSC.}
   \label{fig:req_data}
\end{figure}

\textbf{Number of Additional Unlabelled Data per Task}.
FSC relies on a small amount of task-related unlabeled data to estimate the calibration coefficients. In this paper, we adopt one unlabeled sample per task as the default setting. 
To examine whether the quantity of unlabeled data affects the final performance after calibration, we conduct experiments whose results are presented in Fig.~\ref{fig:req_data}.

The results show that access to task-specific unlabeled data is highly 
beneficial. Even a single additional unlabeled sample per task yields a 
substantial improvement in the performance of the merged model. 
However, as the amount of unlabeled data increases, the performance gains become marginal.

\begin{figure}[htb]
  \centering
   \includegraphics[width=1\linewidth]{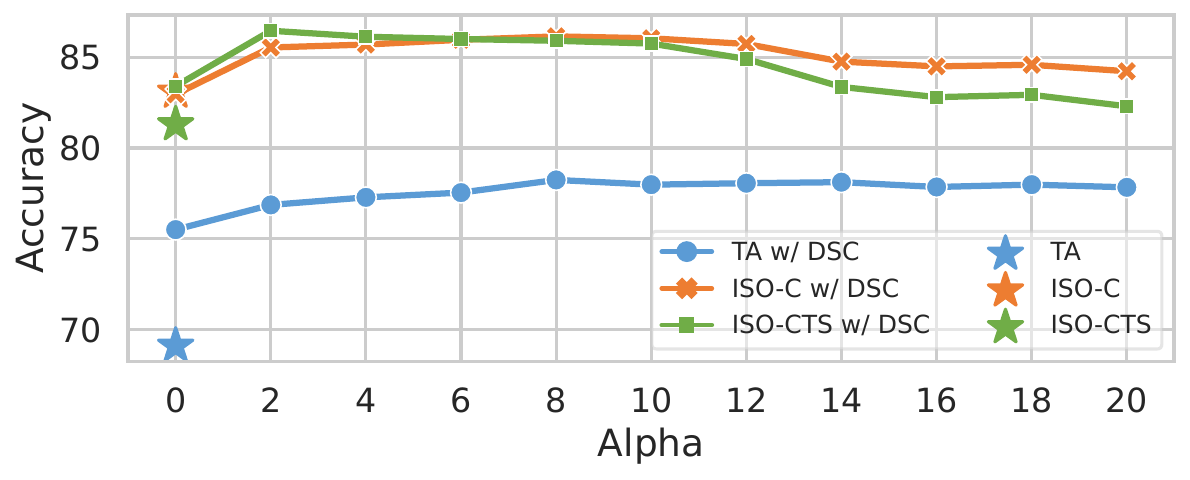}
   \caption{Relations between performance and hyper-parameter.}
   \label{fig:hyper_Alpha_ablation}
\end{figure}

\textbf{Different Hyper-parameter}. 
We conduct ablation studies on the hyperparameter \(\alpha\) in DSC, which 
determines the size of the magnitude-sensitive layer set \(\mathcal{A}\). 
The resulting performance trends are shown in Fig.~\ref{fig:hyper_Alpha_ablation}. 
Two observations can be drawn.

First, the method demonstrates robustness with respect to \(\alpha\). 
Even when \(\alpha=0\), which means that no layer receives differentiated 
treatment, DSC maintains strong performance. 

Second, when magnitude-sensitive layers are included, performance is often 
further improved. This confirms the importance of identifying sensitivity layers.

Throughout the experiments presented in this paper, \(\alpha\) is consistently set to 10. From an empirical perspective, a smaller value of \(\alpha\) is sufficient for any unseen backbone.

\begin{figure}[htb]
  \centering
   \includegraphics[width=1\linewidth]{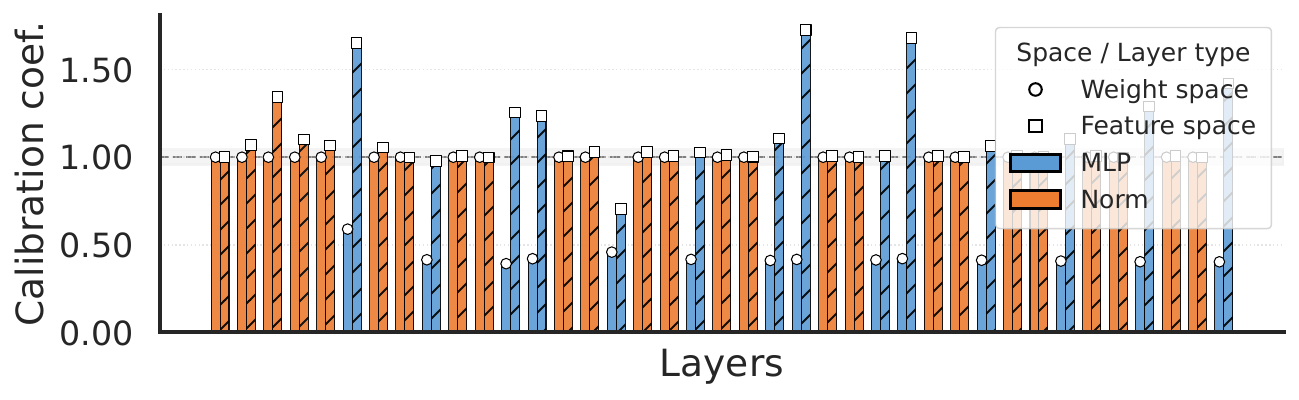}
   \caption{Comparison of calibration coef. For Iso-C between different spaces.}
   \label{fig:ratios_comparison_TIES}
\end{figure}

\textbf{Calibration Coefficients between WSC and FSC}.
Based on Lemma~\ref {lem:tf_alignment}, approximating FSC solely at the weight level neglects the alignment between the task vector and the input-specific Jacobian spectrum, which can occasionally lead to errors in WSC. For instance, Fig.~\ref{fig:ratios_comparison_TIES} visualises the specific calibration coefficients for Iso-C in both feature and weight spaces, and a notable divergence is observed (\textit{i.e.}, feature space requires an increase in magnitude, while weight space demands a reduction). This phenomenon explains the performance degradation observed when applying WSC to Iso-C and Iso-CTS in Tab.~\ref{tab:main_tab2}. Nevertheless, for most merging methods (\textit{e.g.}, TIES, TSV-M), calibration in the weight space remains a reliable approximation.

\begin{figure}[htb]
  \centering
   \includegraphics[width=1\linewidth]{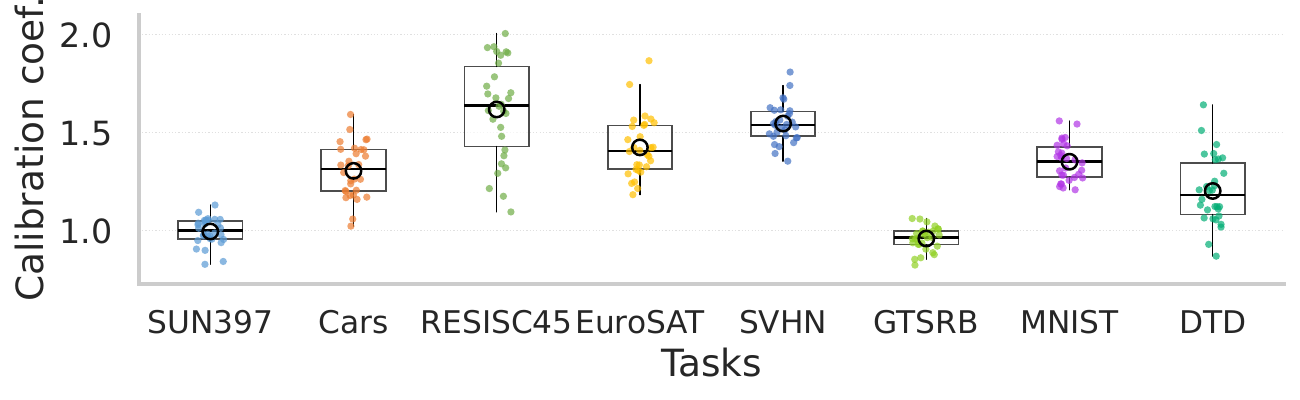}
   \caption{Comparison of calibration coef. For Iso-C between different datasets.}
   \label{fig:coef_across_datasets_box_strip}
\end{figure}

\textbf{Calibration Coefficients between datasets.}
We visualised the estimated calibration coefficients for a certain layer using different datasets in Fig. \ref{fig:coef_across_datasets_box_strip}. It is evident that there are noticeable differences in calibration coefficients between datasets, but the variance from multiple samples within the same dataset is minimal. This also explains why using just a single sample achieves effective calibration, with limited additional benefit from using more samples.

\begin{figure}[htb]
  \centering
   \includegraphics[width=0.9\linewidth]{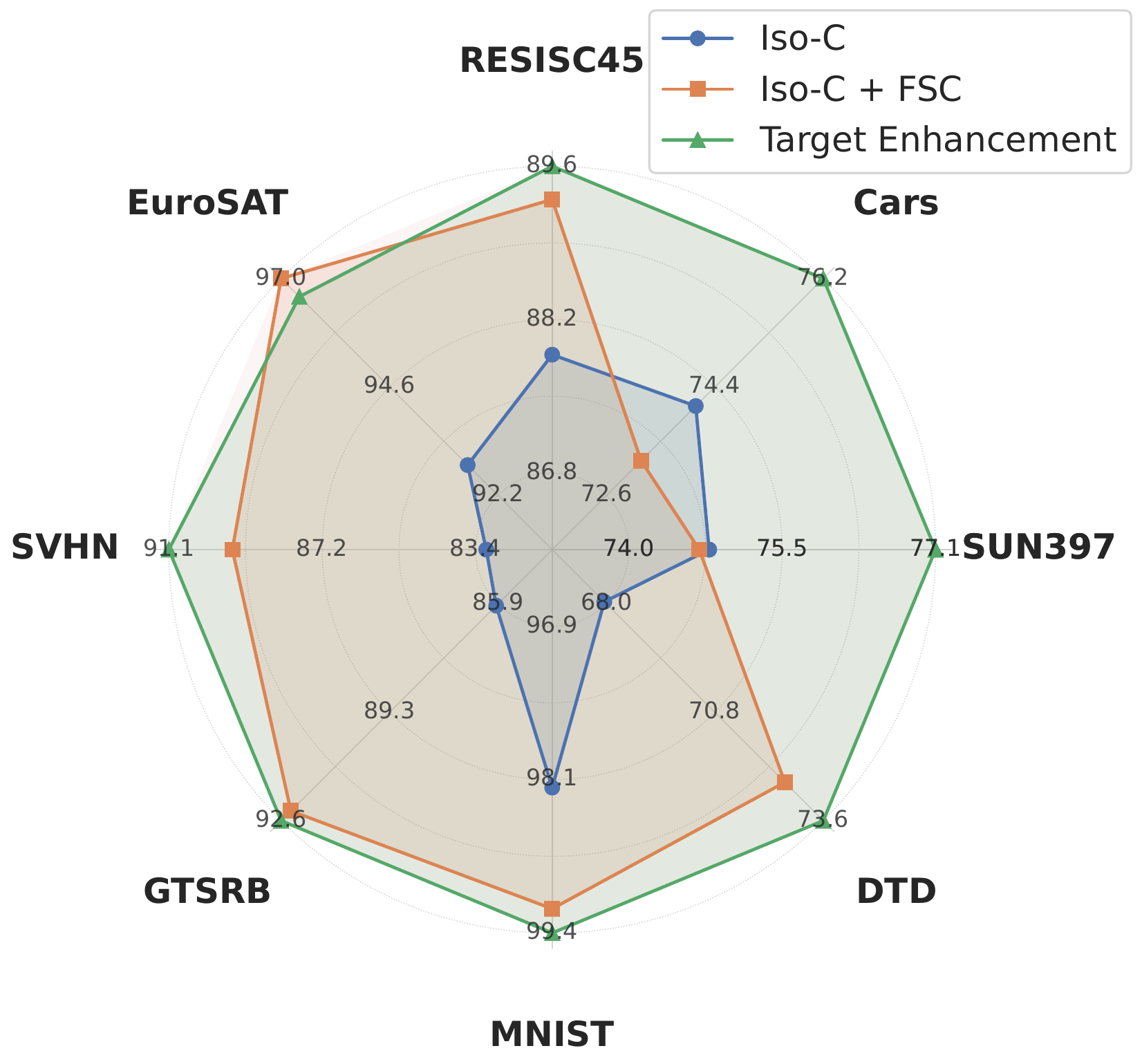}
   \caption{Rescaling the magnitude of the merged model to that of the specialised model can achieve better performance.}
   \label{fig:Target Enhancement}
\end{figure}

\textbf{Enhancement on a specific task}.
Based on Proposition~\ref{prop:optimal_feature_norm}, the merged model achieves optimal task-specific performance when the magnitude of its task feature matches that of the corresponding specialised model (referred to as Target Enhancement). However, in multitask learning scenarios, the magnitude of the merged model’s task feature is rescaled to the average task feature magnitude of the specialised models. This approach improves the merged model’s average performance across multiple tasks, but also reduces the maximal achievable improvement on individual tasks. Take Iso-C as an example, this phenomenon is illustrated in Fig.~\ref{fig:Target Enhancement}.

\begin{figure}[htb]
  \centering
   \includegraphics[width=1\linewidth]{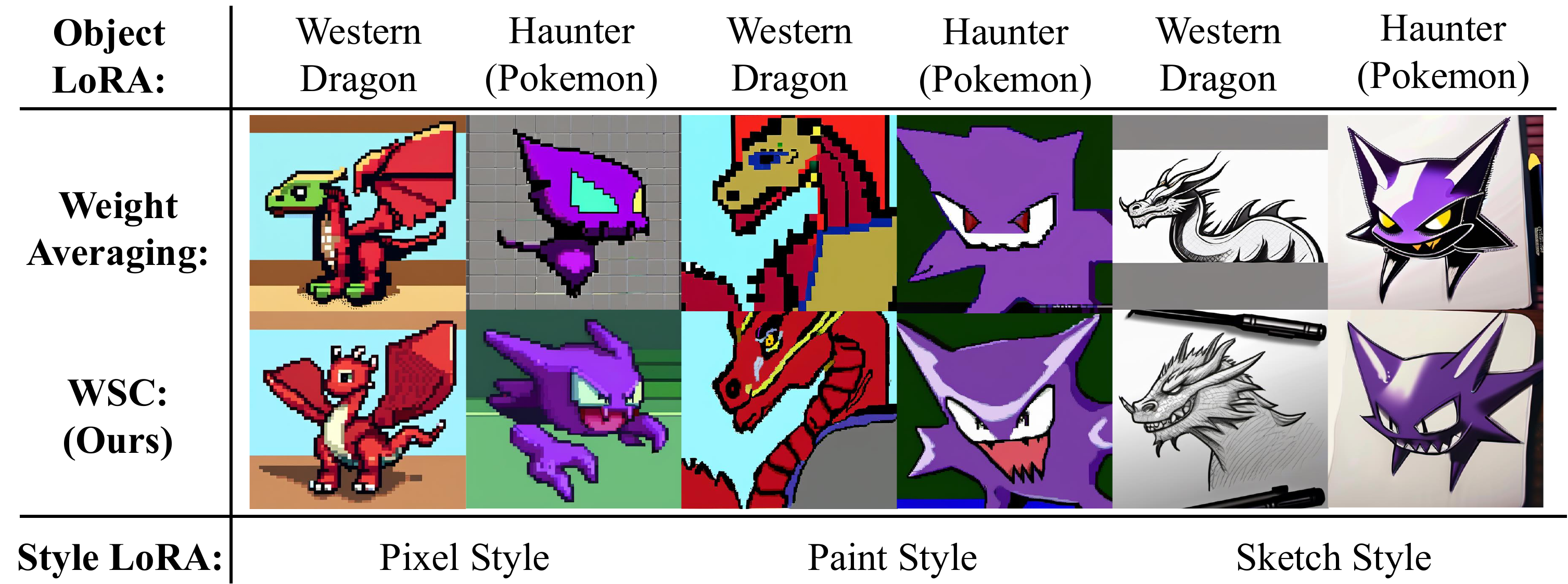}
   \caption{Visualisation of Merging Different Object LoRA and Style LoRA.}
   \label{fig:diffusion}
\end{figure}

\textbf{Merge Diffusion Models}.
Beyond classification tasks, our method enables merging multiple LoRAs \cite{Hu_2021_LoRA} trained on the Stable Diffusion model \cite{Rombach_2022_High-Resolution} for style mixing. Previous work \cite{Shah_2025_ZipLoRA} proposes combining LoRAs trained on specific content (\textit{e.g.}, a particular animal) with those for artistic styles (\textit{e.g.}, painting styles), allowing diffusion models to generate any object-style combination. In this study, we follow this LoRA merging paradigm and download pre-trained LoRA weights from Civitai. We use \textit{Western Dragon} and \textit{Haunter (Pokemon)} as objects, and combine them with \textit{Pixel Style}, \textit{Paint Style}, and \textit{Sketch Style} LoRAs. Visualisation results between Average Weight and its WSC variant are shown in Fig.~\ref{fig:diffusion}.

\textbf{Statistical Significance}.
Due to fixed dataset splits, the main experiment could not yield multiple different results. To better illustrate the significant improvement of our method, we experiment by merging different combinations of models selected from $8$ models, resulting in $(2^8-1)=255$ combinations. By employing TA as the baseline, the final mean accuracies are $77.2$ and $83.0$ before and after calibration, respectively. The t-test of the accuracies between our method and TA yielded a \textit{t-value} of $31.4$ and a \textit{p-value} of $1.5e-89$, confirming our calibration is significant.

\textbf{Task-agnostic dataset for FSC/DSC}.
We recognise that even one piece of task-specific, unlabeled data per task can significantly improve the merged model. Here, we explore using task-agnostic datasets (\textit{e.g.}, ImageNet) as substitutes for minimal task-specific unlabeled data. We term the dual space calibration guided by a task-agnostic dataset as \textit{DSC-A}.

The results in Tab. \ref{tab:cost_abalation} show that due to distributional discrepancies, using a task-agnostic dataset (16 per task) struggles to accurately estimate the magnitude of variation.

\begin{table}[!t]
    \centering
    \caption{Comparison of unlabelled data requirements and time consumption for different methods. Methods with ``-A'' use a task-agnostic dataset to estimate $\xi$ in feature space.}
    \footnotesize
    \begin{tabularx}{\linewidth}{lccC}
        \toprule
        Method & Req. Data (per task) & Time (min) & Acc. (\%) \\
        \midrule
        TA             & 0             & 2    & 69.1  \\
        TA w/ WSC$_\text{ (Ours)}$      & 0             & 2    & 76.6  \\
        TA w/ DSC-A$_\text{ (Ours)}$    & 0             & 2.5  & 76.5  \\
        TA w/ DSC$_\text{ (Ours)}$      & 1             & 2.5  & 78.9  \\
        \bottomrule
    \end{tabularx}
    \label{tab:cost_abalation}
\end{table}


\section{Related Work}
\label{sec:related_work}

\subsection{Model Merging for Multi-Task Learning} 



Compared to Ensemble Learning, which combines predictions from multiple models, model merging integrates models at the parameter level to retain a single model for inference, which can address domain shift \cite{Izmailov_2019_Averaging, Wortsman_2022_Model}, catastrophic forgetting \cite{Chitale_2023_Task, Zhu_2024_Model, Marczak_2024_MagMax}, and construction challenges of large language models (LLMs) \cite{Yang_2024_Model}.
Related methods are categorised based on application timing: 
(i) \textbf{Pre-Merging Methods} which establish favourable conditions for model merging. 
Works include: Ortiz-Jimenez et al.~\cite{Ortiz-Jimenez_2023_Task} proposed linearised fine-tuning of pretrained models in tangent space~\cite{Jacot_2018_Neural}. Jin et al.~\cite{Jin_2024_Fine-Tuning} suggested linear fine-tuning solely on the linear layers within attention modules. OTFusion~\cite{Singh_2020_Model} employs optimal transport to align neurons across different models.
(ii) \textbf{During-Merging Methods}, which focus on optimizing merging strategies; Works include: Task Arithmetic~\cite{Ilharco_2022_Editinga} uses task vectors for merging by computing arithmetic mean. TIES-Merge~\cite{Yadav_2023_TIES-Merging} addresses parameter conflicts and sign discrepancies. DARE~\cite{Yu_2024_Language} reduces parameter interference by setting most task vectors to zero.
(iii) \textbf{Post-Merging Methods}, which adjust the merged models. This is an emerging field. Feature Surgery~\cite{Yang_2024_Representation} utilises unlabelled data to train a ``Surgery'' module, reducing representation bias and enhancing the weight disentanglement property.
Our method falls into Post-Merging Methods that uniquely focus on the adjustment of the model magnitude.

\subsection{Traditional Multi-Task Learning} 

Traditional Multi-Task Learning (MTL) improves learning efficiency and prediction accuracy by jointly training related tasks \cite{Liu_2023_Hierarchical, 9112648, 9568893}. A major challenge is minimising negative transfer \cite{Liu_2021_Conflict-Averse, 5288526}, addressed through architectural design \cite{Sun_2020_AdaShare, Misra_2016_Cross-Stitch} and optimisation dynamics \cite{Sener_2018_Multi-Task, Liu_2021_Conflict-Averse}. 

Architectural methods involve designing networks that share knowledge and retain task-specific layers to prevent interference \cite{Ruder_2017_Overviewa}, using hard parameter sharing \cite{Caruana_1993_Multitask} and mixtures of experts for soft sharing \cite{Ma_2018_Modeling, Hazimeh_2021_DSelect-k, Wang_2022_Multi-Task}. Balancing task contributions by adjusting gradients \cite{Yu_2020_Gradient, Sener_2018_Multi-Task, Chen_2020_Just} and exploring weight assignments strategies \cite{Kendall_2018_Multi-Task, Liu_2019_End-End, Huq_2023_Adaptive}.
In contrast to these traditional approaches, this paper explores efficient MTL through novel model merging methods, which offer higher resource efficiency and are orthogonal to traditional MTL approaches.

\section{Conclusion} 
\label{sec:conclusion}


In this paper, we focus on calibrating the merged model in terms of magnitude.
To achieve this goal, we explore a training-free calibration in feature, weight and dual space, all with minimal computational cost. Our calibration significantly improved the performance of merged models. 
We believe this study advances the understanding of model merging, yet questions remain regarding the assessment of weight disentanglement at the weight level and the realisation of improved calibration under data-free conditions. Future research will continue to address these limitations.


\bibliography{main}

@misc{MNIST,
  title = {{{MNIST}} Handwritten Digit Database, {{Yann LeCun}}, {{Corinna Cortes}} and {{Chris Burges}}},
  howpublished = {https://yann.lecun.com/exdb/mnist/}
}

@inproceedings{Almeida_2011_Contributions,
  title = {Contributions to the Study of {{SMS}} Spam Filtering: New Collection and Results},
  shorttitle = {Contributions to the Study of {{SMS}} Spam Filtering},
  booktitle = {Proceedings of the 11th {{ACM}} Symposium on {{Document}} Engineering},
  author = {Almeida, Tiago A. and Hidalgo, Jos{\'e} Mar{\'i}a G. and Yamakami, Akebo},
  year = {2011},
  month = sep,
  series = {{{DocEng}} '11},
  pages = {259--262},
  publisher = {Association for Computing Machinery},
  address = {New York, NY, USA},
  doi = {10.1145/2034691.2034742},
  isbn = {978-1-4503-0863-2}
}

@inproceedings{Caruana_1993_Multitask,
  title = {Multitask {{Learning}}: {{A Knowledge-Based Source}} of {{Inductive Bias}}},
  shorttitle = {Multitask {{Learning}}},
  booktitle = {Machine {{Learning Proceedings}} 1993},
  author = {Caruana, Richard A.},
  year = {1993},
  pages = {41--48},
  publisher = {Elsevier},
  doi = {10.1016/B978-1-55860-307-3.50012-5},
  copyright = {https://www.elsevier.com/tdm/userlicense/1.0/},
  isbn = {978-1-55860-307-3}
}

@inproceedings{Chen_2020_Just,
  title = {Just {{Pick}} a {{Sign}}: {{Optimizing Deep Multitask Models}} with {{Gradient Sign Dropout}}},
  shorttitle = {Just {{Pick}} a {{Sign}}},
  booktitle = {Advances in {{Neural Information Processing Systems}}},
  author = {Chen, Zhao and Ngiam, Jiquan and Huang, Yanping and Luong, Thang and Kretzschmar, Henrik and Chai, Yuning and Anguelov, Dragomir},
  year = {2020},
  volume = {33},
  pages = {2039--2050},
  publisher = {Curran Associates, Inc.}
}

@article{Cheng_2017_Remote,
  title = {Remote {{Sensing Image Scene Classification}}: {{Benchmark}} and {{State}} of the {{Art}}},
  shorttitle = {Remote {{Sensing Image Scene Classification}}},
  author = {Cheng, Gong and Han, Junwei and Lu, Xiaoqiang},
  year = {2017},
  month = oct,
  journal = {Proceedings of the IEEE},
  volume = {105},
  number = {10},
  pages = {1865--1883},
  issn = {1558-2256},
  doi = {10.1109/JPROC.2017.2675998}
}

@misc{Chitale_2023_Task,
  title = {Task {{Arithmetic}} with {{LoRA}} for {{Continual Learning}}},
  author = {Chitale, Rajas and Vaidya, Ankit and Kane, Aditya and Ghotkar, Archana},
  year = {2023},
  month = nov,
  number = {arXiv:2311.02428},
  eprint = {2311.02428},
  publisher = {arXiv},
  doi = {10.48550/arXiv.2311.02428},
  archiveprefix = {arXiv}
}

@inproceedings{Cimpoi_2014_Describinga,
  title = {Describing {{Textures}} in the {{Wild}}},
  booktitle = {Proceedings of the {{IEEE Conference}} on {{Computer Vision}} and {{Pattern Recognition}}},
  author = {Cimpoi, Mircea and Maji, Subhransu and Kokkinos, Iasonas and Mohamed, Sammy and Vedaldi, Andrea},
  year = {2014},
  pages = {3606--3613}
}

@misc{Gargiulo_2025_Task,
  title = {Task {{Singular Vectors}}: {{Reducing Task Interference}} in {{Model Merging}}},
  shorttitle = {Task {{Singular Vectors}}},
  author = {Gargiulo, Antonio Andrea and Crisostomi, Donato and Bucarelli, Maria Sofia and Scardapane, Simone and Silvestri, Fabrizio and Rodol{\`a}, Emanuele},
  year = {2025},
  month = apr,
  number = {arXiv:2412.00081},
  eprint = {2412.00081},
  primaryclass = {cs},
  publisher = {arXiv},
  doi = {10.48550/arXiv.2412.00081},
  archiveprefix = {arXiv}
}

@misc{Goddard_2024_Arcees,
  title = {Arcee's {{MergeKit}}: {{A Toolkit}} for {{Merging Large Language Models}}},
  shorttitle = {Arcee's {{MergeKit}}},
  author = {Goddard, Charles and Siriwardhana, Shamane and Ehghaghi, Malikeh and Meyers, Luke and Karpukhin, Vlad and Benedict, Brian and McQuade, Mark and Solawetz, Jacob},
  year = {2024},
  month = mar,
  number = {arXiv:2403.13257},
  eprint = {2403.13257},
  publisher = {arXiv},
  doi = {10.48550/arXiv.2403.13257},
  archiveprefix = {arXiv}
}

@inproceedings{Hazimeh_2021_DSelect-k,
  title = {{{DSelect-k}}: {{Differentiable Selection}} in the {{Mixture}} of {{Experts}} with {{Applications}} to {{Multi-Task Learning}}},
  shorttitle = {{{DSelect-k}}},
  booktitle = {Advances in {{Neural Information Processing Systems}}},
  author = {Hazimeh, Hussein and Zhao, Zhe and Chowdhery, Aakanksha and Sathiamoorthy, Maheswaran and Chen, Yihua and Mazumder, Rahul and Hong, Lichan and Chi, Ed},
  year = {2021},
  volume = {34},
  pages = {29335--29347},
  publisher = {Curran Associates, Inc.}
}

@article{Helber_2019_EuroSATa,
  title = {{{EuroSAT}}: {{A Novel Dataset}} and {{Deep Learning Benchmark}} for {{Land Use}} and {{Land Cover Classification}}},
  shorttitle = {{{EuroSAT}}},
  author = {Helber, Patrick and Bischke, Benjamin and Dengel, Andreas and Borth, Damian},
  year = {2019},
  month = jul,
  journal = {IEEE Journal of Selected Topics in Applied Earth Observations and Remote Sensing},
  volume = {12},
  number = {7},
  pages = {2217--2226},
  issn = {2151-1535},
  doi = {10.1109/JSTARS.2019.2918242}
}

@misc{Hu_2021_LoRA,
  title = {{{LoRA}}: {{Low-Rank Adaptation}} of {{Large Language Models}}},
  shorttitle = {{{LoRA}}},
  author = {Hu, Edward J. and Shen, Yelong and Wallis, Phillip and {Allen-Zhu}, Zeyuan and Li, Yuanzhi and Wang, Shean and Wang, Lu and Chen, Weizhu},
  year = {2021},
  month = oct,
  number = {arXiv:2106.09685},
  eprint = {2106.09685},
  publisher = {arXiv},
  doi = {10.48550/arXiv.2106.09685},
  archiveprefix = {arXiv}
}

@misc{Huq_2023_Adaptive,
  title = {Adaptive {{Weight Assignment Scheme For Multi-task Learning}}},
  author = {Huq, Aminul and Pervin, Mst Tasnim},
  year = {2023},
  month = mar,
  number = {arXiv:2303.07278},
  eprint = {2303.07278},
  publisher = {arXiv},
  doi = {10.48550/arXiv.2303.07278},
  archiveprefix = {arXiv}
}

@inproceedings{Ilharco_2022_Editinga,
  title = {Editing Models with Task Arithmetic},
  booktitle = {The {{Eleventh International Conference}} on {{Learning Representations}}},
  author = {Ilharco, Gabriel and Ribeiro, Marco Tulio and Wortsman, Mitchell and Schmidt, Ludwig and Hajishirzi, Hannaneh and Farhadi, Ali},
  year = {2022},
  month = sep
}

@misc{Izmailov_2019_Averaging,
  title = {Averaging {{Weights Leads}} to {{Wider Optima}} and {{Better Generalization}}},
  author = {Izmailov, Pavel and Podoprikhin, Dmitrii and Garipov, Timur and Vetrov, Dmitry and Wilson, Andrew Gordon},
  year = {2019},
  month = feb,
  number = {arXiv:1803.05407},
  eprint = {1803.05407},
  primaryclass = {cs, stat},
  publisher = {arXiv},
  doi = {10.48550/arXiv.1803.05407},
  archiveprefix = {arXiv}
}

@inproceedings{Jacot_2018_Neural,
  title = {Neural {{Tangent Kernel}}: {{Convergence}} and {{Generalization}} in {{Neural Networks}}},
  shorttitle = {Neural {{Tangent Kernel}}},
  booktitle = {Advances in {{Neural Information Processing Systems}}},
  author = {Jacot, Arthur and Gabriel, Franck and Hongler, Clement},
  year = {2018},
  volume = {31},
  publisher = {Curran Associates, Inc.}
}

@misc{Jin_2024_Fine-Tuning,
  title = {Fine-{{Tuning Linear Layers Only Is}} a {{Simple}} yet {{Effective Way}} for {{Task Arithmetic}}},
  author = {Jin, Ruochen and Hou, Bojian and Xiao, Jiancong and Su, Weijie and Shen, Li},
  year = {2024},
  month = jul,
  number = {arXiv:2407.07089},
  eprint = {2407.07089},
  publisher = {arXiv},
  doi = {10.48550/arXiv.2407.07089},
  archiveprefix = {arXiv}
}

@inproceedings{Kendall_2018_Multi-Task,
  title = {Multi-{{Task Learning Using Uncertainty}} to {{Weigh Losses}} for {{Scene Geometry}} and {{Semantics}}},
  booktitle = {Proceedings of the {{IEEE Conference}} on {{Computer Vision}} and {{Pattern Recognition}}},
  author = {Kendall, Alex and Gal, Yarin and Cipolla, Roberto},
  year = {2018},
  pages = {7482--7491}
}

@inproceedings{Krause_2013_3Da,
  title = {{{3D Object Representations}} for {{Fine-Grained Categorization}}},
  booktitle = {Proceedings of the {{IEEE International Conference}} on {{Computer Vision Workshops}}},
  author = {Krause, Jonathan and Stark, Michael and Deng, Jia and {Fei-Fei}, Li},
  year = {2013},
  pages = {554--561}
}

@article{Krizhevsky__Learning,
  title = {Learning {{Multiple Layers}} of {{Features}} from {{Tiny Images}}},
  author = {Krizhevsky, Alex}
}

@inproceedings{Liu_2019_End-End,
  title = {End-{{To-End Multi-Task Learning With Attention}}},
  booktitle = {Proceedings of the {{IEEE}}/{{CVF Conference}} on {{Computer Vision}} and {{Pattern Recognition}}},
  author = {Liu, Shikun and Johns, Edward and Davison, Andrew J.},
  year = {2019},
  pages = {1871--1880}
}

@inproceedings{Liu_2021_Conflict-Averse,
  title = {Conflict-{{Averse Gradient Descent}} for {{Multi-task}} Learning},
  booktitle = {Advances in {{Neural Information Processing Systems}}},
  author = {Liu, Bo and Liu, Xingchao and Jin, Xiaojie and Stone, Peter and Liu, Qiang},
  year = {2021},
  volume = {34},
  pages = {18878--18890},
  publisher = {Curran Associates, Inc.}
}

@inproceedings{Liu_2023_Hierarchical,
  title = {Hierarchical {{Prompt Learning}} for {{Multi-Task Learning}}},
  booktitle = {Proceedings of the {{IEEE}}/{{CVF Conference}} on {{Computer Vision}} and {{Pattern Recognition}}},
  author = {Liu, Yajing and Lu, Yuning and Liu, Hao and An, Yaozu and Xu, Zhuoran and Yao, Zhuokun and Zhang, Baofeng and Xiong, Zhiwei and Gui, Chenguang},
  year = {2023},
  pages = {10888--10898}
}

@inproceedings{Ma_2018_Modeling,
  title = {Modeling {{Task Relationships}} in {{Multi-task Learning}} with {{Multi-gate Mixture-of-Experts}}},
  booktitle = {Proceedings of the 24th {{ACM SIGKDD International Conference}} on {{Knowledge Discovery}} \& {{Data Mining}}},
  author = {Ma, Jiaqi and Zhao, Zhe and Yi, Xinyang and Chen, Jilin and Hong, Lichan and Chi, Ed H.},
  year = {2018},
  month = jul,
  series = {{{KDD}} '18},
  pages = {1930--1939},
  publisher = {Association for Computing Machinery},
  address = {New York, NY, USA},
  doi = {10.1145/3219819.3220007},
  isbn = {978-1-4503-5552-0}
}

@misc{Marczak_2025_No,
  title = {No {{Task Left Behind}}: {{Isotropic Model Merging}} with {{Common}} and {{Task-Specific Subspaces}}},
  shorttitle = {No {{Task Left Behind}}},
  author = {Marczak, Daniel and Magistri, Simone and Cygert, Sebastian and Twardowski, Bart{\l}omiej and Bagdanov, Andrew D. and van de Weijer, Joost},
  year = {2025},
  month = feb,
  number = {arXiv:2502.04959},
  eprint = {2502.04959},
  primaryclass = {cs},
  publisher = {arXiv},
  doi = {10.48550/arXiv.2502.04959},
  archiveprefix = {arXiv}
}

@inproceedings{Misra_2016_Cross-Stitch,
  title = {Cross-{{Stitch Networks}} for {{Multi-Task Learning}}},
  booktitle = {Proceedings of the {{IEEE Conference}} on {{Computer Vision}} and {{Pattern Recognition}}},
  author = {Misra, Ishan and Shrivastava, Abhinav and Gupta, Abhinav and Hebert, Martial},
  year = {2016},
  pages = {3994--4003}
}

@article{Nair_2024_MaxFusion,
  title = {{{MaxFusion}}: {{Plug}}\&amp;{{Play Multi-Modal Generation}} in {{Text-to-Image Diffusion Models}}},
  shorttitle = {{{MaxFusion}}},
  author = {Nair, Nithin Gopalakrishnan and Valanarasu, Jeya Maria Jose and Patel, Vishal M},
  year = {2024},
  publisher = {arXiv},
  doi = {10.48550/ARXIV.2404.09977},
  copyright = {Creative Commons Attribution 4.0 International}
}

@article{Netzer__Reading,
  title = {Reading {{Digits}} in {{Natural Images}} with {{Unsupervised Feature Learning}}},
  author = {Netzer, Yuval and Wang, Tao and Coates, Adam and Bissacco, Alessandro and Wu, Bo and Ng, Andrew Y}
}

@article{Ortiz-Jimenez_2023_Task,
  title = {Task {{Arithmetic}} in the {{Tangent Space}}: {{Improved Editing}} of {{Pre-Trained Models}}},
  shorttitle = {Task {{Arithmetic}} in the {{Tangent Space}}},
  author = {{Ortiz-Jimenez}, Guillermo and Favero, Alessandro and Frossard, Pascal},
  year = {2023},
  month = dec,
  journal = {Advances in Neural Information Processing Systems},
  volume = {36},
  pages = {66727--66754}
}

@misc{Pang_2005_Seeing,
  title = {Seeing Stars: {{Exploiting}} Class Relationships for Sentiment Categorization with Respect to Rating Scales},
  shorttitle = {Seeing Stars},
  author = {Pang, Bo and Lee, Lillian},
  year = {2005},
  month = jun,
  number = {arXiv:cs/0506075},
  eprint = {cs/0506075},
  publisher = {arXiv},
  doi = {10.48550/arXiv.cs/0506075},
  archiveprefix = {arXiv}
}

@inproceedings{Rombach_2022_High-Resolution,
  title = {High-{{Resolution Image Synthesis With Latent Diffusion Models}}},
  booktitle = {Proceedings of the {{IEEE}}/{{CVF Conference}} on {{Computer Vision}} and {{Pattern Recognition}}},
  author = {Rombach, Robin and Blattmann, Andreas and Lorenz, Dominik and Esser, Patrick and Ommer, Bj{\"o}rn},
  year = {2022},
  pages = {10684--10695}
}

@misc{Ruder_2017_Overviewa,
  title = {An {{Overview}} of {{Multi-Task Learning}} in {{Deep Neural Networks}}},
  author = {Ruder, Sebastian},
  year = {2017},
  month = jun,
  number = {arXiv:1706.05098},
  eprint = {1706.05098},
  publisher = {arXiv},
  archiveprefix = {arXiv}
}

@inproceedings{Sener_2018_Multi-Task,
  title = {Multi-{{Task Learning}} as {{Multi-Objective Optimization}}},
  booktitle = {Advances in {{Neural Information Processing Systems}}},
  author = {Sener, Ozan and Koltun, Vladlen},
  year = {2018},
  volume = {31},
  publisher = {Curran Associates, Inc.}
}

@inproceedings{Shah_2025_ZipLoRA,
  title = {{{ZipLoRA}}: {{Any Subject}} in~{{Any Style}} by~{{Effectively Merging LoRAs}}},
  shorttitle = {{{ZipLoRA}}},
  booktitle = {Computer {{Vision}} -- {{ECCV}} 2024},
  author = {Shah, Viraj and Ruiz, Nataniel and Cole, Forrester and Lu, Erika and Lazebnik, Svetlana and Li, Yuanzhen and Jampani, Varun},
  editor = {Leonardis, Ale{\v s} and Ricci, Elisa and Roth, Stefan and Russakovsky, Olga and Sattler, Torsten and Varol, G{\"u}l},
  year = {2025},
  pages = {422--438},
  publisher = {Springer Nature Switzerland},
  address = {Cham},
  doi = {10.1007/978-3-031-73232-4_24},
  isbn = {978-3-031-73232-4}
}

@inproceedings{Singh_2020_Model,
  title = {Model {{Fusion}} via {{Optimal Transport}}},
  booktitle = {Advances in {{Neural Information Processing Systems}}},
  author = {Singh, Sidak Pal and Jaggi, Martin},
  year = {2020},
  volume = {33},
  pages = {22045--22055},
  publisher = {Curran Associates, Inc.}
}

@inproceedings{Stallkamp_2011_German,
  title = {The {{German Traffic Sign Recognition Benchmark}}: {{A}} Multi-Class Classification Competition},
  shorttitle = {The {{German Traffic Sign Recognition Benchmark}}},
  booktitle = {The 2011 {{International Joint Conference}} on {{Neural Networks}}},
  author = {Stallkamp, Johannes and Schlipsing, Marc and Salmen, Jan and Igel, Christian},
  year = {2011},
  month = jul,
  pages = {1453--1460},
  issn = {2161-4407},
  doi = {10.1109/IJCNN.2011.6033395}
}

@inproceedings{Sun_2020_AdaShare,
  title = {{{AdaShare}}: {{Learning What To Share For Efficient Deep Multi-Task Learning}}},
  shorttitle = {{{AdaShare}}},
  booktitle = {Advances in {{Neural Information Processing Systems}}},
  author = {Sun, Ximeng and Panda, Rameswar and Feris, Rogerio and Saenko, Kate},
  year = {2020},
  volume = {33},
  pages = {8728--8740},
  publisher = {Curran Associates, Inc.}
}

@inproceedings{Wang_2022_Multi-Task,
  title = {Multi-{{Task Learning}} with {{Calibrated Mixture}} of {{Insightful Experts}}},
  booktitle = {2022 {{IEEE}} 38th {{International Conference}} on {{Data Engineering}} ({{ICDE}})},
  author = {Wang, Sinan and Li, Yumeng and Li, Hongyan and Zhu, Tanchao and Li, Zhao and Ou, Wenwu},
  year = {2022},
  month = may,
  pages = {3307--3319},
  issn = {2375-026X},
  doi = {10.1109/ICDE53745.2022.00312}
}

@article{Warstadt_2019_Neural,
  title = {Neural {{Network Acceptability Judgments}}},
  author = {Warstadt, Alex and Singh, Amanpreet and Bowman, Samuel R.},
  year = {2019},
  month = sep,
  journal = {Transactions of the Association for Computational Linguistics},
  volume = {7},
  pages = {625--641},
  issn = {2307-387X},
  doi = {10.1162/tacl_a_00290}
}

@inproceedings{Wortsman_2022_Model,
  title = {Model Soups: Averaging Weights of Multiple Fine-Tuned Models Improves Accuracy without Increasing Inference Time},
  shorttitle = {Model Soups},
  booktitle = {Proceedings of the 39th {{International Conference}} on {{Machine Learning}}},
  author = {Wortsman, Mitchell and Ilharco, Gabriel and Gadre, Samir Ya and Roelofs, Rebecca and {Gontijo-Lopes}, Raphael and Morcos, Ari S. and Namkoong, Hongseok and Farhadi, Ali and Carmon, Yair and Kornblith, Simon and Schmidt, Ludwig},
  year = {2022},
  month = jun,
  pages = {23965--23998},
  publisher = {PMLR},
  issn = {2640-3498}
}

@article{Xiao_2016_SUNa,
  title = {{{SUN Database}}: {{Exploring}} a {{Large Collection}} of {{Scene Categories}}},
  shorttitle = {{{SUN Database}}},
  author = {Xiao, Jianxiong and Ehinger, Krista A. and Hays, James and Torralba, Antonio and Oliva, Aude},
  year = {2016},
  month = aug,
  journal = {International Journal of Computer Vision},
  volume = {119},
  number = {1},
  pages = {3--22},
  issn = {0920-5691, 1573-1405},
  doi = {10.1007/s11263-014-0748-y}
}

@inproceedings{Xu_2024_Training-Free,
  title = {Training-{{Free Pretrained Model Merging}}},
  booktitle = {Proceedings of the {{IEEE}}/{{CVF Conference}} on {{Computer Vision}} and {{Pattern Recognition}}},
  author = {Xu, Zhengqi and Yuan, Ke and Wang, Huiqiong and Wang, Yong and Song, Mingli and Song, Jie},
  year = {2024},
  pages = {5915--5925}
}

@article{Yadav_2023_TIES-Merging,
  title = {{{TIES-Merging}}: {{Resolving Interference When Merging Models}}},
  shorttitle = {{{TIES-Merging}}},
  author = {Yadav, Prateek and Tam, Derek and Choshen, Leshem and Raffel, Colin A. and Bansal, Mohit},
  year = {2023},
  month = dec,
  journal = {Advances in Neural Information Processing Systems},
  volume = {36},
  pages = {7093--7115}
}

@article{Yang_2017_Deep,
  title = {Deep {{Learning}} for {{Fixed Model Reuse}}},
  author = {Yang, Yang and Zhan, De-Chuan and Fan, Ying and Jiang, Yuan and Zhou, Zhi-Hua},
  year = {2017},
  month = feb,
  journal = {Proceedings of the AAAI Conference on Artificial Intelligence},
  volume = {31},
  number = {1},
  issn = {2374-3468},
  doi = {10.1609/aaai.v31i1.10855},
  copyright = {Copyright (c)}
}

@misc{Yang_2024_AdaMerging,
  title = {{{AdaMerging}}: {{Adaptive Model Merging}} for {{Multi-Task Learning}}},
  shorttitle = {{{AdaMerging}}},
  author = {Yang, Enneng and Wang, Zhenyi and Shen, Li and Liu, Shiwei and Guo, Guibing and Wang, Xingwei and Tao, Dacheng},
  year = {2024},
  month = may,
  number = {arXiv:2310.02575},
  eprint = {2310.02575},
  primaryclass = {cs},
  publisher = {arXiv},
  doi = {10.48550/arXiv.2310.02575},
  archiveprefix = {arXiv}
}

@misc{Yang_2024_Model,
  title = {Model {{Merging}} in {{LLMs}}, {{MLLMs}}, and {{Beyond}}: {{Methods}}, {{Theories}}, {{Applications}} and {{Opportunities}}},
  shorttitle = {Model {{Merging}} in {{LLMs}}, {{MLLMs}}, and {{Beyond}}},
  author = {Yang, Enneng and Shen, Li and Guo, Guibing and Wang, Xingwei and Cao, Xiaochun and Zhang, Jie and Tao, Dacheng},
  year = {2024},
  month = sep,
  number = {arXiv:2408.07666},
  eprint = {2408.07666},
  primaryclass = {cs},
  publisher = {arXiv},
  doi = {10.48550/arXiv.2408.07666},
  archiveprefix = {arXiv}
}

@misc{Yang_2024_Representation,
  title = {Representation {{Surgery}} for {{Multi-Task Model Merging}}},
  author = {Yang, Enneng and Shen, Li and Wang, Zhenyi and Guo, Guibing and Chen, Xiaojun and Wang, Xingwei and Tao, Dacheng},
  year = {2024},
  month = may,
  number = {arXiv:2402.02705},
  eprint = {2402.02705},
  doi = {10.48550/arXiv.2402.02705},
  archiveprefix = {arXiv}
}

@inproceedings{Yu_2020_Gradient,
  title = {Gradient {{Surgery}} for {{Multi-Task Learning}}},
  booktitle = {Advances in {{Neural Information Processing Systems}}},
  author = {Yu, Tianhe and Kumar, Saurabh and Gupta, Abhishek and Levine, Sergey and Hausman, Karol and Finn, Chelsea},
  year = {2020},
  volume = {33},
  pages = {5824--5836},
  publisher = {Curran Associates, Inc.}
}

@inproceedings{Yu_2024_Language,
  title = {Language {{Models}} Are {{Super Mario}}: {{Absorbing Abilities}} from {{Homologous Models}} as a {{Free Lunch}}},
  shorttitle = {Language {{Models}} Are {{Super Mario}}},
  booktitle = {Forty-First {{International Conference}} on {{Machine Learning}}},
  author = {Yu, Le and Yu, Bowen and Yu, Haiyang and Huang, Fei and Li, Yongbin},
  year = {2024},
  month = jun
}

@inproceedings{Zhang_2015_Character-level,
  title = {Character-Level {{Convolutional Networks}} for {{Text Classification}}},
  booktitle = {Advances in {{Neural Information Processing Systems}}},
  author = {Zhang, Xiang and Zhao, Junbo and LeCun, Yann},
  year = {2015},
  volume = {28},
  publisher = {Curran Associates, Inc.}
}

@inproceedings{Zhu_2024_Model,
  title = {Model {{Tailor}}: {{Mitigating Catastrophic Forgetting}} in {{Multi-modal Large Language Models}}},
  shorttitle = {Model {{Tailor}}},
  booktitle = {Forty-First {{International Conference}} on {{Machine Learning}}},
  author = {Zhu, Didi and Sun, Zhongyisun and Li, Zexi and Shen, Tao and Yan, Ke and Ding, Shouhong and Wu, Chao and Kuang, Kun},
  year = {2024},
  month = jun
}

@misc{alpaca_eval,
  author = {Xuechen Li and Tianyi Zhang and Yann Dubois and Rohan Taori and Ishaan Gulrajani and Carlos Guestrin and Percy Liang and Tatsunori B. Hashimoto },
  title = {AlpacaEval: An Automatic Evaluator of Instruction-following Models},
  year = {2023},
  month = {5},
  publisher = {GitHub},
  journal = {GitHub repository},
  howpublished = {\url{https://github.com/tatsu-lab/alpaca_eval}}
}

@article{cobbe2021gsm8k,
  title={Training Verifiers to Solve Math Word Problems},
  author={Cobbe, Karl and Kosaraju, Vineet and Bavarian, Mohammad and Chen, Mark and Jun, Heewoo and Kaiser, Lukasz and Plappert, Matthias and Tworek, Jerry and Hilton, Jacob and Nakano, Reiichiro and Hesse, Christopher and Schulman, John},
  journal={arXiv preprint arXiv:2110.14168},
  year={2021}
}

@article{touvron2023llama,
  title={Llama 2: Open foundation and fine-tuned chat models},
  author={Touvron, Hugo and Martin, Louis and Stone, Kevin and Albert, Peter and Almahairi, Amjad and Babaei, Yasmine and Bashlykov, Nikolay and Batra, Soumya and Bhargava, Prajjwal and Bhosale, Shruti and others},
  journal={arXiv preprint arXiv:2307.09288},
  year={2023}
}

@article{luo2023wizardmath,
  title={Wizardmath: Empowering mathematical reasoning for large language models via reinforced evol-instruct},
  author={Luo, Haipeng and Sun, Qingfeng and Xu, Can and Zhao, Pu and Lou, Jianguang and Tao, Chongyang and Geng, Xiubo and Lin, Qingwei and Chen, Shifeng and Zhang, Dongmei},
  journal={arXiv preprint arXiv:2308.09583},
  year={2023}
}

@inproceedings{devlin2019bert,
  title={Bert: Pre-training of deep bidirectional transformers for language understanding},
  author={Devlin, Jacob and Chang, Ming-Wei and Lee, Kenton and Toutanova, Kristina},
  booktitle={Proceedings of the 2019 conference of the North American chapter of the association for computational linguistics: human language technologies, volume 1 (long and short papers)},
  pages={4171--4186},
  year={2019}
}

@inproceedings{dagan2005pascal,
  title={The pascal recognising textual entailment challenge},
  author={Dagan, Ido and Glickman, Oren and Magnini, Bernardo},
  booktitle={Machine learning challenges workshop},
  pages={177--190},
  year={2005},
  organization={Springer}
}

@inproceedings{de2019commitmentbank,
  title={The commitmentbank: Investigating projection in naturally occurring discourse},
  author={De Marneffe, Marie-Catherine and Simons, Mandy and Tonhauser, Judith},
  booktitle={proceedings of Sinn und Bedeutung},
  volume={23},
  number={2},
  pages={107--124},
  year={2019}
}

@inproceedings{sakaguchi2020winogrande,
  title={Winogrande: An adversarial winograd schema challenge at scale},
  author={Sakaguchi, Keisuke and Le Bras, Ronan and Bhagavatula, Chandra and Choi, Yejin},
  booktitle={Proceedings of the AAAI Conference on Artificial Intelligence},
  volume={34},
  number={05},
  pages={8732--8740},
  year={2020}
}

@article{pilehvar2018wic,
  title={WiC: the word-in-context dataset for evaluating context-sensitive meaning representations},
  author={Pilehvar, Mohammad Taher and Camacho-Collados, Jose},
  journal={arXiv preprint arXiv:1808.09121},
  year={2018}
}

@article{levesque2012winograd,
  title={The Winograd schema challenge.},
  author={Levesque, Hector J and Davis, Ernest and Morgenstern, Leora},
  journal={KR},
  volume={2012},
  number={13th},
  pages={3},
  year={2012}
}

@inproceedings{roemmele2011choice,
  title={Choice of Plausible Alternatives: An Evaluation of Commonsense Causal Reasoning.},
  author={Roemmele, Melissa and Bejan, Cosmin Adrian and Gordon, Andrew S},
  booktitle={AAAI spring symposium: logical formalizations of commonsense reasoning},
  pages={90--95},
  year={2011}
}

@article{zellers2019hellaswag,
  title={Hellaswag: Can a machine really finish your sentence?},
  author={Zellers, Rowan and Holtzman, Ari and Bisk, Yonatan and Farhadi, Ali and Choi, Yejin},
  journal={arXiv preprint arXiv:1905.07830},
  year={2019}
}

@inproceedings{sharma2018tackling,
  title={Tackling the story ending biases in the story cloze test},
  author={Sharma, Rishi and Allen, James and Bakhshandeh, Omid and Mostafazadeh, Nasrin},
  booktitle={Proceedings of the 56th Annual Meeting of the Association for Computational Linguistics (Volume 2: Short Papers)},
  pages={752--757},
  year={2018}
}

@article{nie2019adversarial,
  title={Adversarial NLI: A new benchmark for natural language understanding},
  author={Nie, Yixin and Williams, Adina and Dinan, Emily and Bansal, Mohit and Weston, Jason and Kiela, Douwe},
  journal={arXiv preprint arXiv:1910.14599},
  year={2019}
}

@String(ECCV= {Eur. Conf. Comput. Vis.})

@String(AAAI = {AAAI})

@String(ECCV  = {ECCV})

@misc{Ainsworth_2023_Git,
  title = {Git {{Re-Basin}}: {{Merging Models}} modulo {{Permutation Symmetries}}},
  shorttitle = {Git {{Re-Basin}}},
  author = {Ainsworth, Samuel K. and Hayase, Jonathan and Srinivasa, Siddhartha},
  year = {2023},
  month = mar,
  number = {arXiv:2209.04836},
  eprint = {2209.04836},
  primaryclass = {cs},
  publisher = {arXiv},
  doi = {10.48550/arXiv.2209.04836},
  archiveprefix = {arXiv}
}

@misc{Marczak_2024_MagMax,
  title = {{{MagMax}}: {{Leveraging Model Merging}} for {{Seamless Continual Learning}}},
  shorttitle = {{{MagMax}}},
  author = {Marczak, Daniel and Twardowski, Bart{\l}omiej and Trzci{\'n}ski, Tomasz and Cygert, Sebastian},
  year = {2024},
  month = jul,
  number = {arXiv:2407.06322},
  eprint = {2407.06322},
  publisher = {arXiv},
  doi = {10.48550/arXiv.2407.06322},
  archiveprefix = {arXiv}
}

@inproceedings{Shoemake_1985_Animating,
  title = {Animating Rotation with Quaternion Curves},
  booktitle = {Proceedings of the 12th Annual Conference on {{Computer}} Graphics and Interactive Techniques},
  author = {Shoemake, Ken},
  year = {1985},
  month = jul,
  series = {{{SIGGRAPH}} '85},
  pages = {245--254},
  publisher = {Association for Computing Machinery},
  address = {New York, NY, USA},
  doi = {10.1145/325334.325242},
  isbn = {978-0-89791-166-5}
}

@misc{Gargiulo_2024_Task,
  title = {Task {{Singular Vectors}}: {{Reducing Task Interference}} in {{Model Merging}}},
  shorttitle = {Task {{Singular Vectors}}},
  author = {Gargiulo, Antonio Andrea and Crisostomi, Donato and Bucarelli, Maria Sofia and Scardapane, Simone and Silvestri, Fabrizio and Rodol{\`a}, Emanuele},
  year = {2024},
  month = nov,
  number = {arXiv:2412.00081},
  eprint = {2412.00081},
  primaryclass = {cs},
  publisher = {arXiv},
  doi = {10.48550/arXiv.2412.00081},
  archiveprefix = {arXiv}
}

@misc{Du_2025_Neural,
  title = {Neural {{Parameter Search}} for {{Slimmer Fine-Tuned Models}} and {{Better Transfer}}},
  author = {Du, Guodong and Fang, Zitao and Li, Jing and Li, Junlin and Jiang, Runhua and Yu, Shuyang and Guo, Yifei and Chen, Yangneng and Goh, Sim Kuan and Tang, Ho-Kin and He, Daojing and Liu, Honghai and Zhang, Min},
  year = {2025},
  month = may,
  number = {arXiv:2505.18713},
  eprint = {2505.18713},
  primaryclass = {cs},
  publisher = {arXiv},
  doi = {10.48550/arXiv.2505.18713},
  archiveprefix = {arXiv}
}

@article{tang2023parameter,
  title={Parameter efficient multi-task model fusion with partial linearization},
  author={Tang, Anke and Shen, Li and Luo, Yong and Zhan, Yibing and Hu, Han and Du, Bo and Chen, Yixin and Tao, Dacheng},
  journal={arXiv preprint arXiv:2310.04742},
  year={2023}
}

@inproceedings{NEURIPS2018_5a4be1fa,
 author = {Jacot, Arthur and Gabriel, Franck and Hongler, Clement},
 booktitle = {Advances in Neural Information Processing Systems},
 editor = {S. Bengio and H. Wallach and H. Larochelle and K. Grauman and N. Cesa-Bianchi and R. Garnett},
 pages = {},
 publisher = {Curran Associates, Inc.},
 title = {Neural Tangent Kernel: Convergence and Generalization in Neural Networks},
 url = {https://proceedings.neurips.cc/paper_files/paper/2018/file/5a4be1fa34e62bb8a6ec6b91d2462f5a-Paper.pdf},
 volume = {31},
 year = {2018}
}

@inproceedings{radford2021learning,
  title={Learning transferable visual models from natural language supervision},
  author={Radford, Alec and Kim, Jong Wook and Hallacy, Chris and Ramesh, Aditya and Goh, Gabriel and Agarwal, Sandhini and Sastry, Girish and Askell, Amanda and Mishkin, Pamela and Clark, Jack and others},
  booktitle={International conference on machine learning},
  pages={8748--8763},
  year={2021},
  organization={PmLR}
}

@ARTICLE{5288526,
  author={Pan, Sinno Jialin and Yang, Qiang},
  journal={IEEE Transactions on Knowledge and Data Engineering}, 
  title={A Survey on Transfer Learning}, 
  year={2010},
  volume={22},
  number={10},
  pages={1345-1359},
  doi={10.1109/TKDE.2009.191}}

@ARTICLE{9447164,
  author={Wu, Xi-Zhu and Xu, Wenkai and Liu, Song and Zhou, Zhi-Hua},
  journal={IEEE Transactions on Knowledge and Data Engineering}, 
  title={Model Reuse With Reduced Kernel Mean Embedding Specification}, 
  year={2023},
  volume={35},
  number={1},
  pages={699-710},
  doi={10.1109/TKDE.2021.3086619}}

@ARTICLE{9112648,
  author={Gui, Lin and Leng, Jia and Zhou, Jiyun and Xu, Ruifeng and He, Yulan},
  journal={IEEE Transactions on Knowledge and Data Engineering}, 
  title={Multi Task Mutual Learning for Joint Sentiment Classification and Topic Detection}, 
  year={2022},
  volume={34},
  number={4},
  pages={1915-1927},
  doi={10.1109/TKDE.2020.2999489}}

@ARTICLE{9568893,
  author={Zhu, Fujin and Lu, Jie and Lin, Adi and Xuan, Junyu and Zhang, Guangquan},
  journal={IEEE Transactions on Knowledge and Data Engineering}, 
  title={Direct Learning With Multi-Task Neural Networks for Treatment Effect Estimation}, 
  year={2023},
  volume={35},
  number={3},
  pages={2457-2470},
  doi={10.1109/TKDE.2021.3112591}}
\bibliographystyle{IEEEtran}

\end{document}